\documentclass[lettersize,journal]{IEEEtran}
\usepackage{amsmath,amsfonts}
\usepackage{algorithmic}
\usepackage{algorithm}
\usepackage{array}
\usepackage[caption=false,font=normalsize,labelfont=sf,textfont=sf]{subfig}
\usepackage{textcomp}
\usepackage{stfloats}
\usepackage{url}
\usepackage{verbatim}
\usepackage{graphicx}
\usepackage{cite}

\usepackage{multirow}
\usepackage{bbding}
\usepackage{colortbl}
\usepackage{makecell}
\usepackage{booktabs}
\usepackage[dvipsnames,svgnames]{xcolor}

\usepackage{hyperref}  

\begin{document}

\title{Towards Robust Multimodal Emotion Recognition under Missing Modalities and Distribution Shifts}

\author{Guowei~Zhong,~Ruohong Huan,~Mingzhen~Wu,~Ronghua~Liang,~\IEEEmembership{Senior Member,~IEEE}, and Peng~Chen,~\IEEEmembership{Member,~IEEE}
\thanks{This work is supported by the National Natural Science Foundation of China (grant number 62276237, 62036009, 62432014), Basic Public Welfare Research Program of Zhejiang Province (LTGY23F020006), and Zhejiang Provincial Natural Science Foundation of China (LDT23F0202, LDT23F02021F02).}
\thanks{Guowei Zhong, Ruohong Huan, Mingzhen Wu, and Peng Chen are with the College of Computer Science and Technology, Zhejiang University of Technology, Hangzhou, 310023, China. (E-mail: guoweizhong@zjut.edu.cn, huanrh@zjut.edu.cn, wumingzhen@zjut.edu.cn, chenpeng@zjut.edu.cn.)}
\thanks{Ronghua Liang is with the College of Computer Science and Technology, Zhejiang University of Technology, Hangzhou, 310023, China, and the School of Computer Science and Technology, Zhejiang University of Science and Technology, Hangzhou, 310023, China. (E-mail: rhliang@zjut.edu.cn)}
\thanks{This work has been submitted to the IEEE for possible publication. Copyright may be transferred without notice, after which this version may no longer be accessible.}}

\markboth{Journal of \LaTeX\ Class Files,~Vol.~14, No.~8, August~2021}%
{Shell \MakeLowercase{\textit{et al.}}: A Sample Article Using IEEEtran.cls for IEEE Journals}


\maketitle

\begin{abstract}
Recent advancements in Multimodal Emotion Recognition (MER) face challenges in addressing both modality missing and Out-Of-Distribution (OOD) data simultaneously. Existing methods often rely on specific models or introduce excessive parameters, which limits their practicality. To address these issues, we propose a novel robust MER framework, \underline{C}ausal \underline{I}nference \underline{D}istill\underline{er} (CIDer), and introduce a new task, Random Modality Feature Missing (RMFM), to generalize the definition of modality missing. CIDer integrates two key components: a Model-Specific Self-Distillation (MSSD) module and a Model-Agnostic Causal Inference (MACI) module. MSSD enhances robustness under the RMFM task through a weight-sharing self-distillation approach applied across low-level features, attention maps, and high-level representations. Additionally, a Word-level Self-aligned Attention Module (WSAM) reduces computational complexity, while a Multimodal Composite Transformer (MCT) facilitates efficient multimodal fusion. To tackle OOD challenges, MACI employs a tailored causal graph to mitigate label and language biases using a Multimodal Causal Module (MCM) and fine-grained counterfactual texts. Notably, MACI can independently enhance OOD generalization with minimal additional parameters. Furthermore, we also introduce the new repartitioned MER OOD datasets. Experimental results demonstrate that CIDer achieves robust performance in both RMFM and OOD scenarios, with fewer parameters and faster training compared to state-of-the-art methods. The implementation of this work is publicly accessible at \url{https://github.com/gw-zhong/CIDer}.
\end{abstract}


\section{Introduction}
\IEEEPARstart{I}{n} real-world scenarios, emotions are conveyed through multiple modalities, including language, pitch, facial expressions, and body movements. Multimodal Emotion Recognition (MER) leverages these modalities to interpret emotional states. While earlier research primarily focused on fusing word-level aligned multimodal sequences~\cite{zadeh2017tensor, liu2018efficient, wang2019words, huan2024trisat, ouyang2024distinguishing}, recent advancements aim to apply MER in practical, real-world contexts, addressing challenges such as modality missing and out-of-distribution (OOD)\footnote{In this paper, to align with the prior work~\cite{sun2022counterfactual}, the described OOD pertains specifically to the deviations in word distribution and the resulting deviations in label distribution caused by such word distribution shifts.} generalization.

\begin{figure}[!t]
	\centering
	\includegraphics[width=\linewidth]{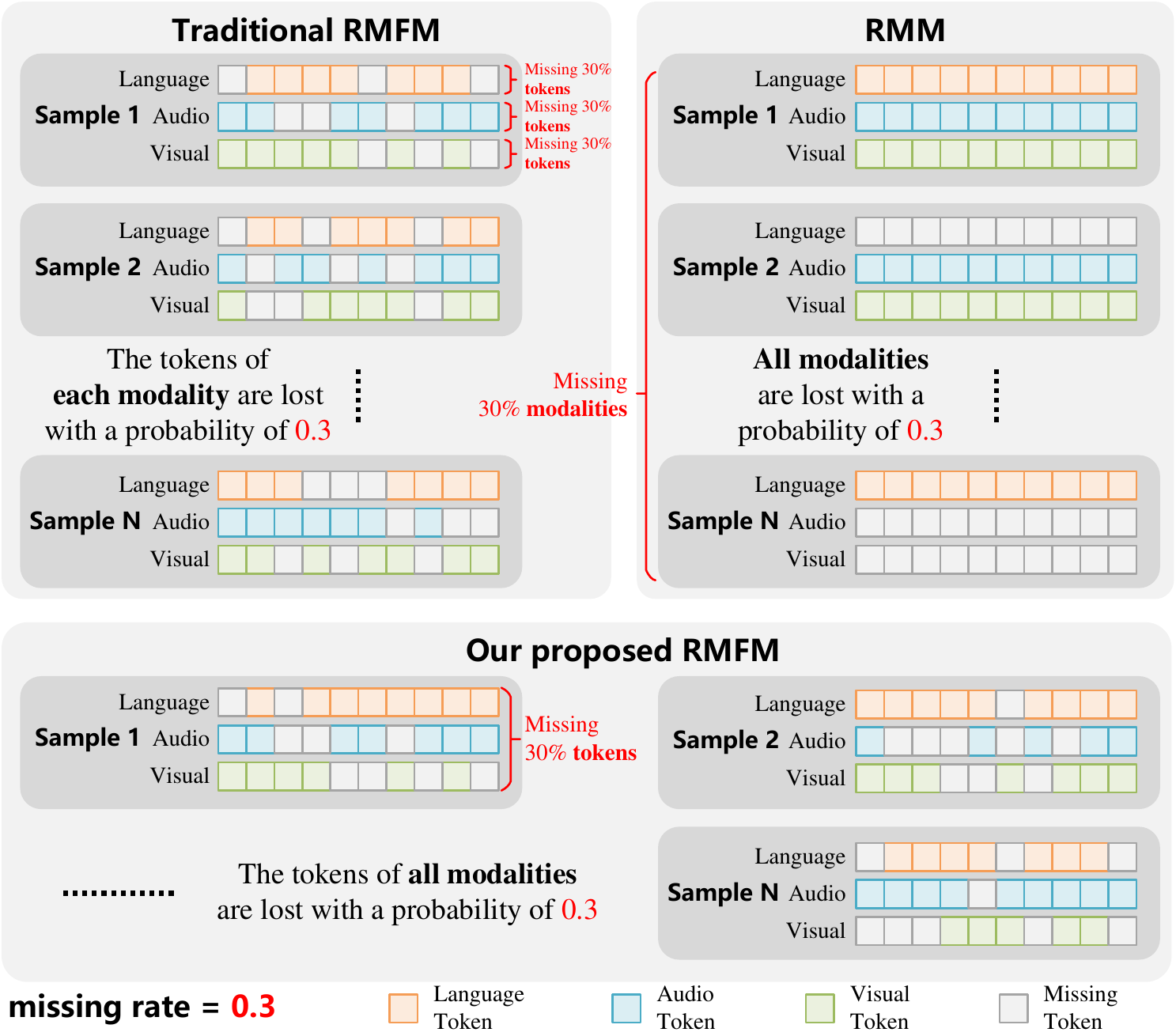}
	\caption{Definition of modality missing. 1) Traditional RMFM: each feature within a modality has a 30\% probability of being lost. 2) RMM: a 30\% probability of an entire modality being lost across the dataset. 3) Our Proposed RMFM: a 30\% probability of losing any feature across all modalities, providing a more generalized approach to modality missing.}
	\label{task_des}
\end{figure}

To address modality missing in MER, some researchers employ modality translation to infer missing data~\cite{pham2019found, wang2020transmodality, tang2021ctfn, huan2023unimf}, while others reconstruct missing features to guide learning without requiring complete multimodal sequences~\cite{yuan2021transformer, sun2023efficient, lian2023gcnet}. However, these approaches have notable limitations: 1) They often assume that training and test data are independently and identically distributed (IID), resulting in poor performance on OOD data. 2) Their definitions of modality missing are inadequate. Current methods typically handle two missing scenarios: missing features across modalities with equal probability (Traditional Random Modality Feature Missing, Traditional RMFM) or random missing modalities across samples (Random Modality Missing, RMM). These definitions fail to account for more complex cases, such as partial frame loss in video clips while retaining language and audio data. Neither traditional RMFM nor RMM can adequately describe such scenarios, as illustrated in Fig.~\ref{task_des}.

To address OOD generalization in MER, researchers have employed causal inference to mitigate bias~\cite{sun2022counterfactual, yang2024towards, huan2024muldef} and utilized probabilistic or model-based approaches~\cite{sun2023general, ma2024bcd} to enhance generalization. However, these methods exhibit notable limitations: 1) They often perform poorly when modalities are missing. 2) Many require specific debiasing models, complicating the enhancement of OOD generalization in existing MER frameworks. Even model-agnostic methods tend to introduce excessive learnable parameters, increasing computational overhead. 3) Current word-based MER OOD datasets present critical flaws: they provide only word-level alignment, mix OOD and IID data in test sets (thereby reducing the effective OOD sample size and increasing variance), and frequently include mixed sentences, which limits their utility.

To address these challenges, we propose \textbf{C}ausal \textbf{I}nference \textbf{D}istill\textbf{er} (CIDer), a robust MER framework that effectively handles both modality missing and OOD issues. Additionally, we introduce a novel RMFM task, which generalizes the concept of modality missing by using a missing rate to describe the loss of features across all three modalities, encompassing traditional RMFM and RMM as special cases. To address RMFM, we propose a Model-Specific Self-Distillation (MSSD) module. For OOD challenges, we introduce a Model-Agnostic Causal Inference (MACI) module, which introduces minimal learnable parameters and can be independently applied to enhance OOD generalization in existing MER models. Furthermore, we repartitioned the original IID datasets to create OOD datasets with training, validation, and test sample counts almost consistent with the IID dataset, providing both word-level aligned and unaligned versions to facilitate future research.

In summary, the contributions of this paper are as follows:
\begin{itemize}
	\item We introduce CIDer, a robust MER framework that addresses both modality missing and OOD challenges.
	\item We propose a generalized RMFM task and the MSSD module to address it. MSSD employs a hierarchical structure for knowledge self-distillation within a weight-sharing twin network. To reduce computational complexity, we introduce the Word-level Self-aligned Attention Module (WSAM) for word-level alignment of non-linguistic sequences. Additionally, we propose the Multimodal Composite Transformer (MCT) to facilitate efficient multimodal fusion through intra- and inter-modal interactions using shared attention matrices.
	\item We propose a tailored causal graph and the MACI module to address the OOD challenge. Within MACI, we introduce the Multimodal Causal Module (MCM) to mitigate label bias during training and construct fine-grained counterfactual texts to reduce language bias during testing. MACI introduces minimal learnable parameters and can be independently applied to enhance OOD generalization in existing MER models. Furthermore, we repartitioned the IID datasets to create new OOD datasets for MER.
	\item Extensive experiments on two MER datasets demonstrate that CIDer achieves superior robustness in addressing RMFM and OOD challenges, with fewer parameters and faster training compared to state-of-the-art methods.
\end{itemize}
\section{Related Work}
\subsection{Incomplete Data}
\textbf{Traditional Random Modality Feature Missing (Traditional RMFM):} This research area focuses on how models should address scenarios where each modality feature in the input data is randomly missing with equal probability. Yuan et al.~\cite{yuan2021transformer} introduced a feature reconstruction network to enable the model to extract as much semantic information as possible from incomplete features. Sun et al.~\cite{sun2023efficient} proposed the Efficient Multimodal Transformer (EMT), which facilitates interactions between global multimodal and local unimodal representations. Additionally, they introduced Dual-Level Feature Restoration (DLFR) to reconstruct features from a low-dimensional space, encouraging the model to learn semantic information from incomplete data. Yuan et al.~\cite{yuan2023noise} developed a noise-hint-based adversarial training framework to enhance the model’s robustness against various potential defects during inference. Furthermore, Yuan et al.~\cite{yuan2024meta} proposed the Meta Noise Adaptation (Meta-NA) strategy, enabling robust training in the presence of noisy instances. Zhang et al.~\cite{zhang2024towards} introduced the Language-dominated Noise-resistant Learning Network (LNLN) to achieve robust MER. The LNLN incorporates two key components: the Dominant Modality Correction (DMC) module and the Dominant Modality-based Multimodal Learning (DMML) module. These modules improve the model’s robustness across diverse noisy scenarios by preserving the quality of dominant modality representations.

Although this line of research has addressed the traditional RMFM problem to some extent, most studies rely on the assumption that features are missing with equal probability across modalities — an assumption that may not hold true in real-world scenarios.

\textbf{Random Modality Missing (RMM):} This research area focuses on addressing how models should handle scenarios where each modality in the input data is randomly missing with a certain probability. Zeng et al.~\cite{zeng2022tag, zeng2022robust} introduced a method to tackle the issue of RMM by leveraging tags to assist in different modality-missing situations. Furthermore, Zeng et al.~\cite{zeng2022mitigating} proposed a backbone encoder-decoder network to learn joint representations of available modalities and assess semantic consistency to determine whether the missing modality is critical for overall MER. Lian et al.~\cite{lian2023gcnet} introduced two graph neural networks (GNNs) — Speaker GNN and Temporal GNN — which aim to jointly optimize classification and reconstruction tasks to fully utilize both complete and incomplete data in an end-to-end manner. Wang et al.~\cite{wang2023distribution}, to alleviate potential distribution inconsistency between recovered and real data, proposed transferring the distribution of available modalities to missing modalities to preserve distribution consistency in the recovered data. Lin et al.~\cite{lin2023missmodal} employed constraints such as geometric contrastive loss, distribution distance loss, and sentiment semantic loss to align the representations of modality-missing and modality-complete data, thereby enhancing the model's robustness to missing modalities without compromising the recognition of complete modalities. Liu et al.~\cite{liu2024modality}, to fully leverage the effectiveness of the language modality in MER, proposed a Modality Translation-based Multimodal Sentiment Analysis model (MTMSA), which exhibits robustness in uncertain modality-missing situations. Guo et al.~\cite{guo2024multimodal}, addressing the RMM problem, introduced three types of prompts: generation prompts, missing signal prompts, and missing type prompts. These prompts facilitate the generation of missing modality features and promote learning within and across modalities.

Although this line of research has addressed the RMM problem, there may also be real-world scenarios where a sequence of a specific modality is partially missing — a situation referred to as RMFM. In such cases, these approaches often fail to perform effectively.

In summary, although the aforementioned works have enhanced the models' robustness in scenarios involving missing modalities, their performance significantly deteriorates when faced with OOD data. Furthermore, the definitions of modality missing in these two types of works exhibit certain limitations and fail to encompass all possible scenarios of modality absence.
\subsection{Out-of-distribution Data}
In this category, researchers have primarily focused on enhancing the robustness of models when handling OOD data inputs. Sun et al.~\cite{sun2022counterfactual} were the first to introduce the OOD Multimodal Sentiment Analysis (MSA) task and construct a corresponding word-based OOD dataset. By employing counterfactual reasoning to model the text modality independently, they derived single-modality predictions for text. During the testing phase, subtracting the text-only prediction from the multimodal prediction yields a debiased prediction result. Furthermore, Sun et al.~\cite{sun2023general} investigated biases introduced by non-text modalities and proposed a general debiasing framework based on inverse probability weighting, which adaptively assigns lower weights to samples with greater biases. Ma et al.~\cite{ma2024bcd} introduced the Bilateral Cross-modal Debias Multimodal sentiment analysis Model (BCD-MM), which improves the model's generalization ability in OOD scenarios by enhancing the extraction of low-redundancy cross-modal features and reducing reliance on non-causal associations. Yang et al.~\cite{yang2024towards} proposed a causal-based Multimodal Counterfactual Inference Sentiment (MCIS) analysis framework, which imagines two counterfactual scenarios during the reasoning phase to mitigate utterance-level label bias and word-level context bias. Huan et al.~\cite{huan2024muldef} achieved debiasing from both multimodal bias and label bias perspectives by leveraging front-door adjustment and counterfactual reasoning methods.

Although the aforementioned works have enhanced the generalization of models for OOD data, these methods often fall short in addressing modality-missing scenarios. Furthermore, such approaches typically necessitate the construction of specialized models to achieve debiasing, which is highly inconvenient for improving the OOD generalization of existing MER models. Although some model-agnostic approaches have emerged, they still require the introduction of a significant number of additional learnable parameters. Additionally, current word-based MER OOD datasets remain limited by several notable shortcomings.

\section{Methodology}
\begin{figure*}[!t]
	\centering
	\includegraphics[width=\linewidth]{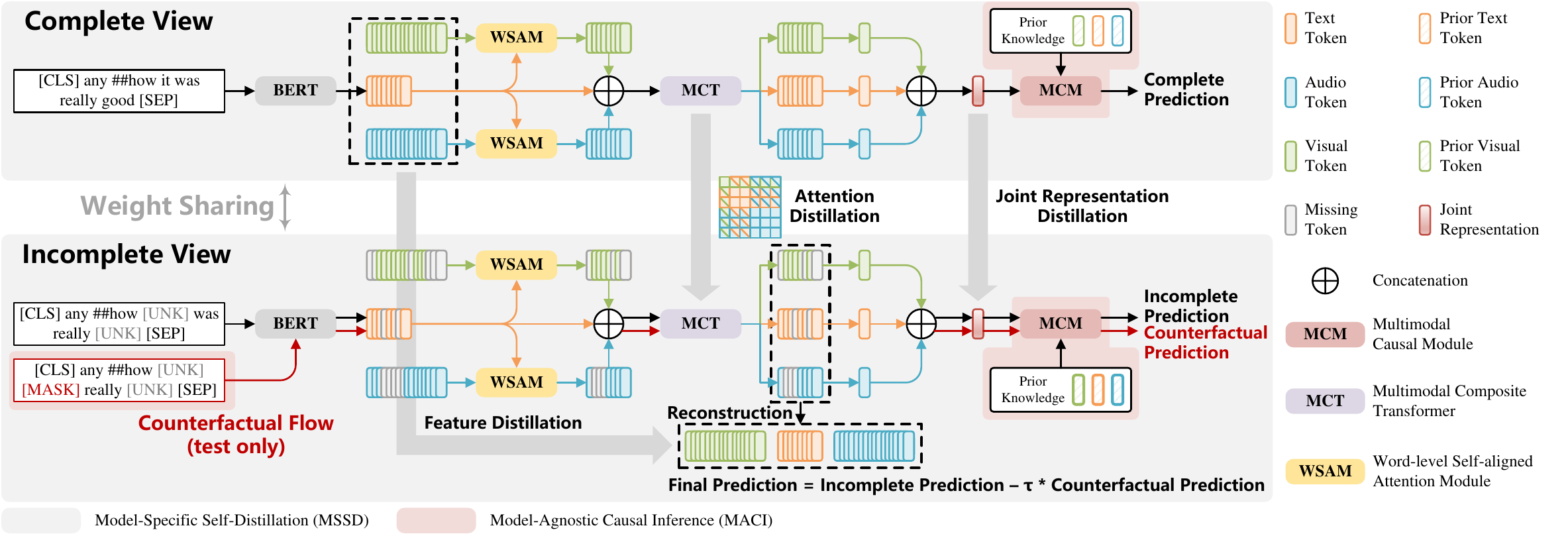}
	\caption{The overall framework of CIDer. Specifically, CIDer consists of two primary modules: the MSSD module and the MACI module. These modules are designed to enhance the model's robustness against incomplete data and improve its generalization capability for OOD data, respectively.}
	\label{framework}
\end{figure*}
In this section, we provide a detailed overview of the proposed CIDer framework. The overall architecture of CIDer is illustrated in Fig.~\ref{framework}.
\subsection{Preliminaries}
\textbf{Data Preparation:} MER typically involves three modalities: text ($\mathring{S} \in \mathbb{R}^{T_l \times 1}$), audio ($\mathring{A} \in \mathbb{R}^{T_a \times d_a}$), and vision ($\mathring{V} \in \mathbb{R}^{T_v \times d_v}$), where $T_{\{l,a,v\}}$ denote the sequence lengths and $d_{\{a,v\}}$ represent the feature dimensions.

For incomplete data, we define: 
\begin{equation}
	(\widetilde{S}, \widetilde{A}, \widetilde{V}) = \text{Missing}((\mathring{S}, \mathring{A}, \mathring{V}))
\end{equation}
where $\text{Missing}(\cdot)$ applies modality missing methods such as our proposed RMFM, Traditional RMFM, RMM, Temporal Modality Feature Missing (TMFM), Structural Temporal Modality Feature Missing (STMFM)~\cite{yuan2023noise} or Specific Modality Missing (SMM). 

For our RMFM, given a missing rate $r\%$, we generate a total missing mask $M_{\text{all}} \in \mathbb{R}^{T_l + T_a + T_v}$, which is split into $M_L \in \mathbb{R}^{T_l}$, $M_A \in \mathbb{R}^{T_a}$, and $M_V \in \mathbb{R}^{T_v}$. $M_k^{(i)} = 0$ ($k \in \{L, A, V\}$) indicates that the $i$-th feature in modality $k$ is missing. The implementation details for other cases of missing modalities can be found in the supplementary materials and the open-source code.

During training, we apply a random missing rate within $[0\%, 100\%)$ to robustly handle various scenarios.

\textbf{Task Definition:} MER aims to leverage information from $(\mathring{S}, \mathring{A}, \mathring{V})$ or $(\widetilde{S}, \widetilde{A}, \widetilde{V})$ to identify shared modality expressions. For classification, the objective is to predict the emotion $y' \in \{1, \ldots, \text{cls}\}$, where $\text{cls}$ represents the total number of categories.

For simplicity, the absence of the $\mathring{}$ or $\widetilde{}$ indicators in the following text implies that both complete and incomplete data are handled in the same manner.
\subsection{Model-Specific Self-Distillation}
\subsubsection{Unimodal Compression}
\begin{figure}[!t]
	\centering
	\includegraphics[width=\linewidth]{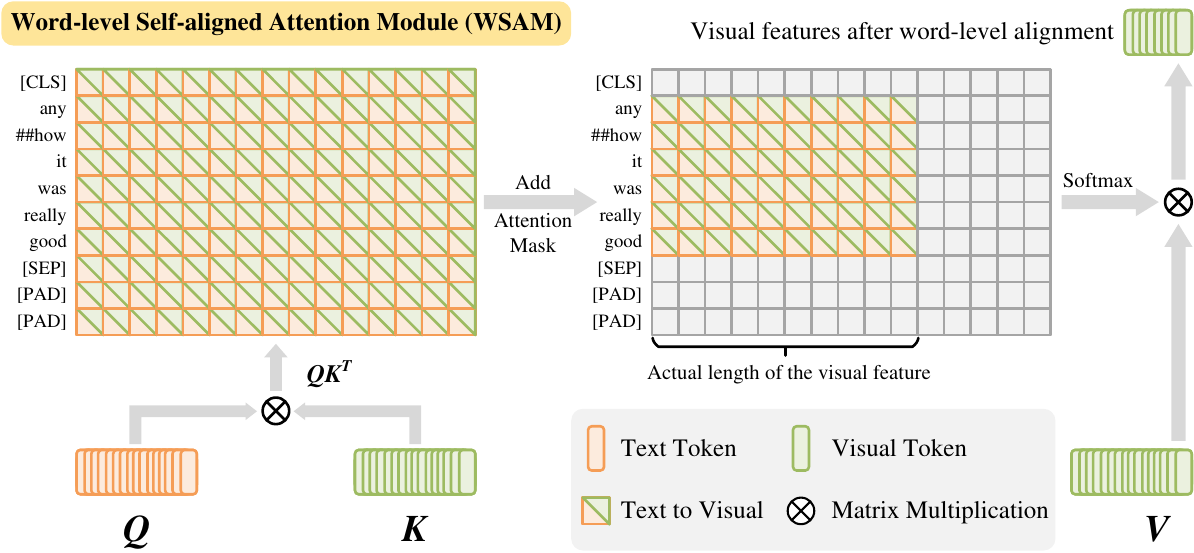}
	\caption{The Word-level Self-aligned Attention Module (WSAM), taking the alignment of the visual sequence to the language sequence as an example.}
	\label{wsam}
\end{figure}
For the text sequence $S$, we first extract features using BERT~\cite{devlin2019bert}:  
\begin{equation}
	L = \text{BERT}\left(S\right) \in \mathbb{R}^{T_l \times d_l}
\end{equation}
where $d_l$ denotes the dimension of the text features. To facilitate subsequent attention computations, we map each modality sequence into a shared feature space using 1D convolution:
\begin{equation}
	X_{\{l,a,v\}} = \text{Conv1D}\left(\left\{L, A, V\right\}\right) \in \mathbb{R}^{T_{\left\{l,a,v\right\}} \times d}
\end{equation}
where $d$ represents the common feature dimension across modalities. To reduce computational complexity and training time, we propose the Word-level Self-aligned Attention Module (WSAM), which aligns long sequences $X_a$ and $X_v$ at the word level.

\textbf{Word-level Self-aligned Attention Module:} As illustrated in Fig.~\ref{wsam}, for the visual sequence $X_v$, WSAM uses $X_l$ as the \texttt{Query} and $X_v$ as the \texttt{Key} and \texttt{Value}, performing alignment through attention:
\begin{equation}
	\begin{aligned}
		X_v' &= \text{WSAM}\left(X_l, X_v\right) \\
		&= \text{softmax}\left(\frac{X_l W_Q W_K^T X_v^T}{\sqrt{d}} + \text{MASK}_v\right) X_v W_V
	\end{aligned}
\end{equation} 
where $W_{\{Q,K,V\}} \in \mathbb{R}^{d \times d}$ are learnable weight matrices, and $\text{MASK}_v \in \mathbb{R}^{T_l \times T_v}$ is used to mask irrelevant tokens. Similarly, the same method aligns the audio sequence $X_a$, yielding $X_a' = \text{WSAM}(X_l, X_a) \in \mathbb{R}^{T_l \times d}$.

Finally, a single-layer GRU~\cite{chung2014empirical} extracts temporal features for each modality:
\begin{equation}
	\begin{aligned}
		H_l &= \text{GRU}\left(X_l\right) \in \mathbb{R}^{T_l \times d} \\
		H_{\{a,v\}} &= \text{GRU}\left(X_{\left\{a,v\right\}}'\right) \in \mathbb{R}^{T_l \times d}
	\end{aligned}
\end{equation}

\subsubsection{Multimodal Fusion}
\begin{figure}[!t]
	\centering
	\includegraphics[width=\linewidth]{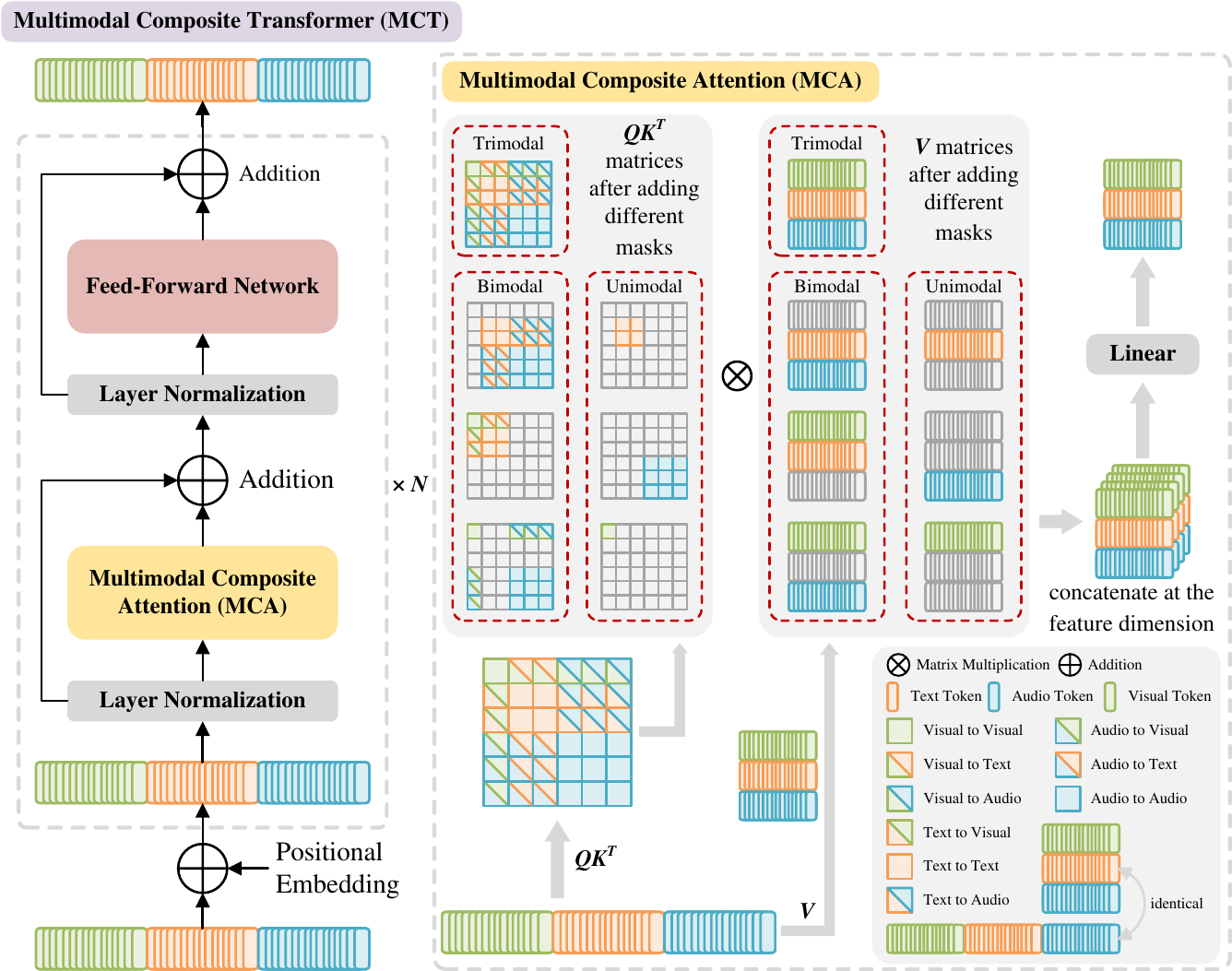}
	\caption{The Multimodal Composite Transformer (MCT), which is primarily composed of Multimodal Composite Attention (MCA).}
	\label{mct}
\end{figure}
After obtaining the temporal features $H_{\{l,a,v\}}$ for each modality through intra-modal interaction, the next step involves inter-modal interaction. First, the temporal features of the three modalities are concatenated along the time dimension:
\begin{equation}
	H_m = \text{concat}\left(H_l, H_a, H_v\right) \in \mathbb{R}^{\left(3 \times T_l\right) \times d}
\end{equation}

The concatenated multimodal sequence $H_m$ is then fed into the Multimodal Composite Transformer (MCT), whose core component is the Multimodal Composite Attention (MCA), as illustrated in Fig.~\ref{mct}.

\textbf{Multimodal Composite Transformer:} The MCA matrix is computed as:  
\begin{equation}
	\text{Attn}_{\text{tri}} = \frac{H_m W_Q W_K^T H_m^T}{\sqrt{d}} \in \mathbb{R}^{\left(3 \times T_l\right) \times \left(3 \times T_l\right)}
\end{equation}

Six mask matrices are applied to $\text{Attn}_{\text{tri}}$ to form unimodal and bimodal interactions, yielding $\text{Attn}_{\text{bi}} \in \{\text{Attn}_{l \leftrightarrow a}, \text{Attn}_{l \leftrightarrow v}, \text{Attn}_{a \leftrightarrow v}\}$ and $\text{Attn}_{\text{uni}} \in \{\text{Attn}_l, \text{Attn}_a, \text{Attn}_v\}$. After applying softmax, these attention matrices are multiplied with $H_{\left\{\text{tri,bi,uni}\right\}} W_V$\footnote{$ H_{\text{tri}} = H_m $, $ H_{\text{bi}} \in \{H_{l \leftrightarrow a}, H_{l \leftrightarrow v}, H_{a \leftrightarrow v}\} $, and $ H_{\text{uni}} \in \{H_l, H_a, H_v\} $. Among these, $ H_{\text{bi}} $ and $ H_{\text{uni}} $ represent subsets of $ H_{\text{tri}} $.} to produce seven multimodal representations:
\begin{equation}
	H'_{\left\{\text{tri,bi,uni}\right\}} = \text{softmax}\left(\text{Attn}_{\left\{\text{tri,bi,uni}\right\}}\right) H_{\left\{\text{tri,bi,uni}\right\}} W_V
\end{equation} 

These representations are mapped through modality-specific linear layers:  
\begin{equation}
	\bar{H}_{\left\{\text{tri,bi,uni}\right\}} = \text{Linear}_{\left\{\text{tri,bi,uni}\right\}}\left(H'_{\left\{\text{tri,bi,uni}\right\}}\right)
\end{equation}

For each modality, the valid parts of $\bar{H}_{\{\text{tri,bi,uni}\}}$ are concatenated along the feature dimension:
\begin{equation}
	\begin{aligned}
		\bar{H}_l &= \text{concat}\left(\bar{H}_{\text{tri}}^l, \bar{H}_{l \leftrightarrow a}^l, \bar{H}_{l \leftrightarrow v}^l, \bar{H}_l^l\right) \in \mathbb{R}^{T_l \times \left(4 \times d\right)} \\
		\bar{H}_a &= \text{concat}\left(\bar{H}_{\text{tri}}^a, \bar{H}_{l \leftrightarrow a}^a, \bar{H}_{a \leftrightarrow v}^a, \bar{H}_a^a\right) \in \mathbb{R}^{T_l \times \left(4 \times d\right)} \\
		\bar{H}_v &= \text{concat}\left(\bar{H}_{\text{tri}}^v, \bar{H}_{l \leftrightarrow v}^v, \bar{H}_{a \leftrightarrow v}^v, \bar{H}_v^v\right) \in \mathbb{R}^{T_l \times \left(4 \times d\right)} \\
	\end{aligned}
\end{equation}
where $\bar{H}_{\text{tri}}^{k_{lav}}$ represents the portion of $\bar{H}_{\text{tri}}$ associated with modality $k_{lav}$ ($k_{lav} \in \{l, a, v\}$), while $\bar{H}_{l \leftrightarrow a}^{k_{la}}$, $\bar{H}_{l \leftrightarrow v}^{k_{lv}}$, and $\bar{H}_{a \leftrightarrow v}^{k_{av}}$ respectively denote the portions of $\bar{H}_{l \leftrightarrow a}$, $\bar{H}_{l \leftrightarrow v}$, and $\bar{H}_{a \leftrightarrow v}$ corresponding to modality $k_{\{la, lv, av\}}$ (with $k_{la} \in \{l, a\}$, $k_{lv} \in \{l, v\}$, and $k_{av} \in \{a, v\}$, respectively). Additionally, $\bar{H}_l^l$, $\bar{H}_a^a$, and $\bar{H}_v^v$ represent the entirety of $\bar{H}_l$, $\bar{H}_a$, and $\bar{H}_v$, respectively, as they pertain to their respective modalities.
These are then reduced in dimension through linear layers:
\begin{equation}
	\bar{\bar{H}}_{\left\{l,a,v\right\}} = \text{Linear}_{\left\{l,a,v\right\}}(\bar{H}_{\left\{l,a,v\right\}}) \in \mathbb{R}^{T_l \times d}
\end{equation}  

Finally, $\bar{\bar{H}}_{\{l,a,v\}}$ are concatenated and passed through a linear layer to obtain the MCA output:
\begin{equation}
	\widehat{H}_m = \text{Linear}\left(\text{concat}\left(\bar{\bar{H}}_l, \bar{\bar{H}}_a, \bar{\bar{H}}_v\right)\right) \in \mathbb{R}^{\left(3 \times T_l\right) \times d}
\end{equation}

For the $i$-th layer of MCT ($i = 1, \ldots, N$), the process is defined as follows:
\begin{equation}
	\begin{aligned}
		H_m^{\left[0\right]} &= H_m \\
		\widehat{H}_m^{\left[i\right]} &= \text{MCA}\left(\text{LN}\left(H_m^{\left[i-1\right]}\right)\right) + \text{LN}\left(H_m^{\left[i-1\right]}\right) \\
		H_m^{\left[i\right]} &= \text{FFN}\left(\text{LN}\left(\widehat{H}_m^{\left[i\right]}\right)\right) + \text{LN}\left(\widehat{H}_m^{\left[i\right]}\right)
	\end{aligned}
\end{equation}
where LN denotes layer normalization~\cite{ba2016layer}, and FFN represents a feed-forward network. After $N$ layers, $H_m^{[N]} \in \mathbb{R}^{(3 \times T_l) \times d}$ is obtained. Next, we separately extract each modality component to obtain $ H_{\{l,a,v\}}^{[N]} \in \mathbb{R}^{T_l \times d} $. The utterance-level representation for each modality is computed using naive attention (NA)~\cite{sun2023efficient}:
\begin{equation}
	G_{\left\{l,a,v\right\}} = \text{NA}\left(H_{\{l,a,v\}}^{\left[N\right]}\right) \in \mathbb{R}^d
\end{equation}

These representations are concatenated and passed through two linear layers to produce the final multimodal joint representation:
\begin{equation}
	h_m = \text{Linear}\left(\text{ReLU}\left(\text{Linear}\left(\text{concat}\left(G_l, G_a, G_v\right)\right)\right)\right) \in \mathbb{R}^{3 \times d}
\end{equation}

\subsubsection{Hierarchical Distillation} 
To enhance the model's robustness in modality-missing scenarios without increasing the training load or compromising its lightweight design, we employ weight-sharing twin networks to perform hierarchical self-distillation across three dimensions—low-level features, attention maps, and high-level joint representations—in a bottom-up manner.

\textbf{Feature Distillation:} First, we reconstruct the outputs of the MCT, $\widetilde{H}_{\{l,a,v\}}^{[N]}$, under incomplete data inputs:
\begin{equation}
	\begin{aligned}
		L_{\text{fake}} &= \text{Linear}\left(\widetilde{H}_l^{[N]}\right) \in \mathbb{R}^{T_l \times d_l} \\
		A_{\text{fake}} &= \text{Linear}\left(\text{ReLU}\left(\text{Linear}\left(\widetilde{H}_a^{\left[N\right]}\right)\right)\right) \in \mathbb{R}^{T_a \times d_a} \\
		V_{\text{fake}} &= \text{Linear}\left(\text{ReLU}\left(\text{Linear}\left(\widetilde{H}_ v^{\left[N\right]}\right)\right)\right) \in \mathbb{R}^{T_v \times d_v}
	\end{aligned}
\end{equation}

We then use SmoothL1 loss to align $\{L_{\text{fake}}, A_{\text{fake}}, V_{\text{fake}}\}$ with the original features $\{\mathring{L}, \mathring{A}, \mathring{V}\}$: 
\begin{equation}
	\begin{aligned}
		L_{\text{recon}} &= \text{SmoothL1}\left(L_\text{fake}, \mathring{L}\right) \\
		& + \text{SmoothL1}\left(A_\text{fake}, \mathring{A}\right) \\
		& + \text{SmoothL1}\left(V_\text{fake}, \mathring{V}\right)
	\end{aligned}
\end{equation}

\textbf{Attention Distillation:} We extract the attention matrices from the last MCT layer under complete $\mathring{\text{Attn}}_{\text{tri}}^{[N]}$ and incomplete $\widetilde{\text{Attn}}_{\text{tri}}^{[N]}$ data inputs. KL divergence is used to align $\widetilde{\text{Attn}}_{\text{tri}}^{[N]}$ with $\mathring{\text{Attn}}_{\text{tri}}^{[N]}$:
\begin{equation}
	L_{\text{attn}} = \text{KL}\left(\widetilde{\text{Attn}}_{\text{tri}}^{\left[N\right]}, \mathring{\text{Attn}}_{\text{tri}}^{\left[N\right]}\right)
\end{equation}

\textbf{Joint Representation Distillation:} We extract the multimodal joint representations under complete $\mathring{h}_m$ and incomplete $\widetilde{h}_m$ data inputs. Cosine similarity is used to align $\widetilde{h}_m$ with $\mathring{h}_m$: 
\begin{equation}
	L_{\text{joint}} = 1 - \text{CosineSimilarity}\left(\widetilde{h}_m, \mathring{h}_m\right)
\end{equation}

\subsection{Model-Agnostic Causal Inference}
\begin{figure}[!t]
	\centering
	\includegraphics[width=\linewidth]{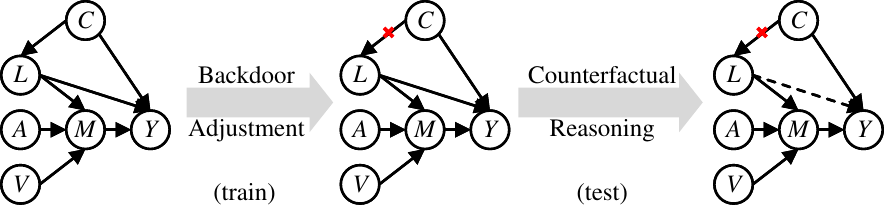}
	\caption{The causal graph for MER. During training, label bias is mitigated through backdoor adjustment, which interrupts the causal relationship $C \rightarrow L$. During testing, language bias is alleviated through counterfactual reasoning, which disrupts the causal relationship $L \rightarrow Y$.}
	\label{causal_graph}
\end{figure}
\begin{figure}[!t]
	\centering
	\includegraphics[width=\linewidth]{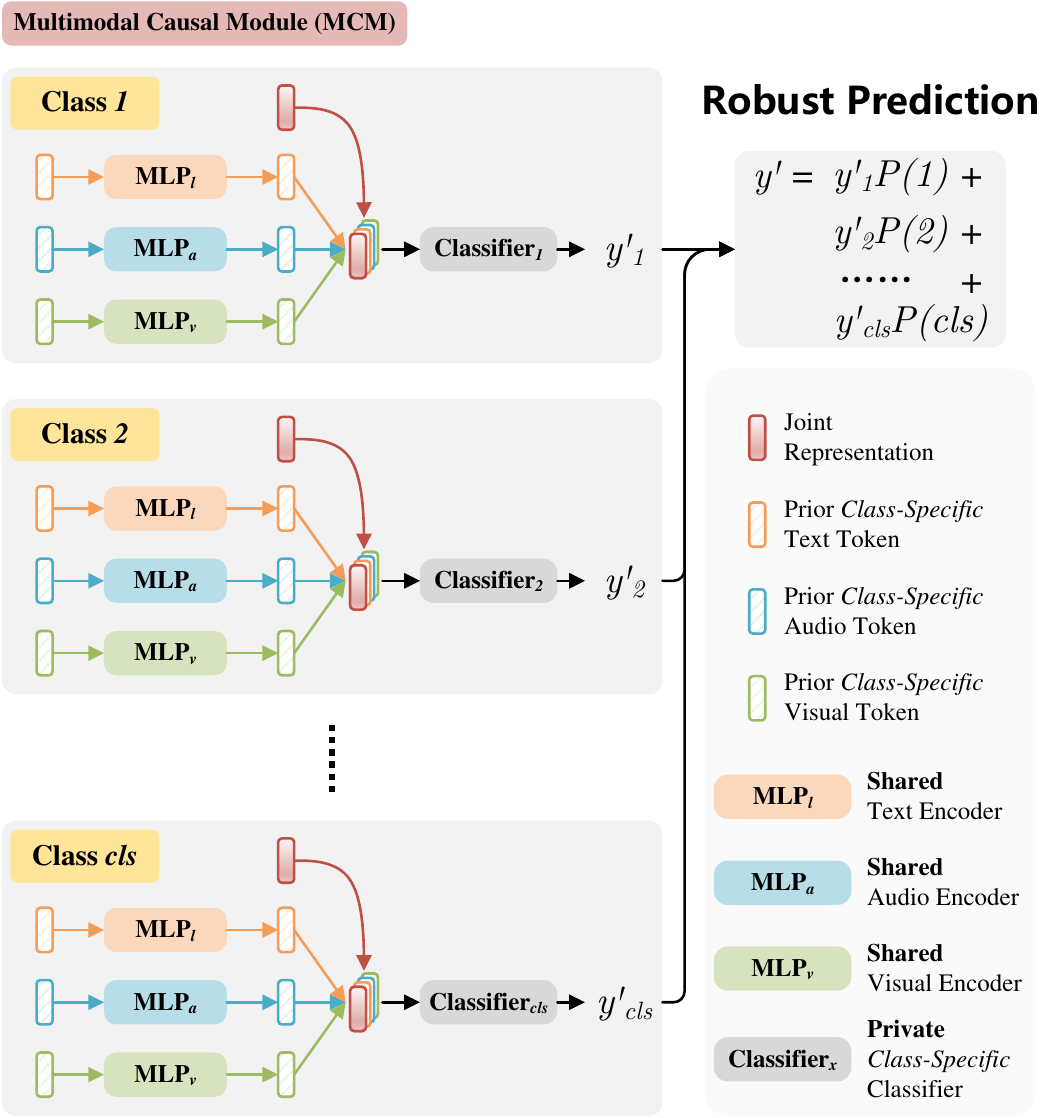}
	\caption{The Multimodal Causal Module (MCM).}
	\label{mcm}
\end{figure}
The imbalance in the distribution of word categories often leads to an implicit imbalance in the distribution of category labels. To achieve robust predictions, we design a novel causal graph for MER that addresses both label bias and language bias, as illustrated in Fig.~\ref{causal_graph}. In the graph, $L$, $A$, and $V$ represent language, audio, and visual inputs, respectively; $M$ denotes the fused multimodal representation; $Y$ represents the label; and $C$ indicates label bias.

\textbf{Label Bias Mitigation:} During training, we employ backdoor adjustment to mitigate label bias. The traditional prediction $P(Y | L) = P(Y | M = f(L, A, V))$ is adjusted to the interventional form: 
\begin{equation}
	P\left(Y | \text{do}\left(L\right)\right) = \sum_{c} P\left(Y | M = f\left(L, A, V\right), c\right) P\left(c\right)
\end{equation}

This is implemented through the Multimodal Causal Module (MCM), as shown in Fig.~\ref{mcm}. For each class $i$ ($i \in \{1, \dots, cls\}$), we compute the mean features $L_i$\footnote{For the language sequence $S_i$, the corresponding feature $L_i$ is obtained using the pre-trained BERT$_\text{BASE}$ model.}, $A_i$, and $V_i$, and encode them using modality-shared MLPs:
\begin{equation}
	H_{i}^{\{{l,a,v}\}} = \text{MLP}_{\{{l,a,v}\}}\left( \left\{ {L,A,V} \right\}_{i} \right) \in \mathbb{R}^{d}
\end{equation}
where $\text{MLP}_{\{l,a,v\}}(\cdot)$ consists of three linear layers interspersed with two ReLU activation. Then, we concatenate them with the multimodal joint representation $h_m$ to form class-specific causal representations $h_i^c$:
\begin{equation}
	h_{i}^{c} = \text{concat}\left( {h_{m},H_{i}^{l},H_{i}^{a},H_{i}^{v}} \right) \in \mathbb{R}^{6 \times d}
\end{equation}

Next, $h_i^c$ is passed through a class-specific classifier (a fully connected layer) to produce the classification result $y_i'$ for category $i$, corresponding to the term $P(Y | M = f(L, A, V), c)$:
\begin{equation}
	y_i' = \text{Classifier}_{i}\left( h_{i}^{c} \right) \in \mathbb{R}^{cls}
\end{equation}

Subsequently, it is multiplied by the pre-calculated prior probability $P(i)$ of each category's label distribution and summed to produce the robust classification output\footnote{During testing, since the category labels are not available, we assume an equal number of instances across all categories. Based on this assumption, we compute the corresponding $\{L, A, V\}_i$ and $P(i)$.}:
\begin{equation}
	y' = {\sum\limits_{i = 1}^{cls}y_{i}'}P(i) \in \mathbb{R}^{cls}
\end{equation}

\textbf{Language Bias Mitigation:} During testing, we use counterfactual reasoning to mitigate language bias. Instead of relying on additional unimodal modeling, we directly adjust $P(Y | \text{do}(L))$ by subtracting counterfactual predictions:
\begin{equation}
	y_{\text{final}}' = P\left(Y | \text{do}\left(L = l\right)\right) - \tau P\left(Y | \text{do}\left(L = l_{\text{cf}}\right)\right)
\end{equation}
where $l$ represents the full text feature, $l_{\text{cf}}$ denotes the counterfactual text feature ($l_{\text{cf}} \subseteq l$), and $\tau$ controls the debiasing strength.

Next, we will detail the process of constructing the fine-grained counterfactual text $S_{\text{cf}}$. First, we compute the frequency of each word $w$ within each class in the dataset to obtain the set $N_w$:
\begin{equation}
	N_w = \{N_w^1, \ldots, N_w^{cls}\}
\end{equation}

Next, we compute the coefficient of variation for the word $w$ using the set $N_w$:
\begin{equation}
	\text{CV}_{w} = \frac{\sigma\left( N_{w} \right)}{\mu\left( N_{w} \right)}
\end{equation}
where $\sigma(\cdot)$ represents the standard deviation, and $\mu(\cdot)$ represents the mean.

We define that when $\text{CV}_w \geq 0.1$, the word $w$ exhibits a significant inter-class distribution difference and is therefore retained in the counterfactual text $S_{\text{cf}}$. To prevent words with uneven inter-class distributions but low overall frequency in the dataset from being categorized as counterfactual, we restrict to the top 100 most frequent words for constructing $S_{\text{cf}}$. For words with $\text{CV}_w < 0.1$ or those with low frequency, we replace them with the \texttt{[MASK]} token.

This results in counterfactual texts such as $S_{\text{cf}} = \langle w_1, \texttt{[MASK]}, \texttt{[MASK]}, \ldots, w_{T_l} \rangle$, which are processed through BERT to obtain $l_{\text{cf}}$ and reduce language bias.

\subsection{Two-stage Optimization}  
To fully leverage information from both complete and incomplete modalities during training, we adopt a two-stage optimization approach. First, the model weights are updated using complete modalities through a conventional classification task:
\begin{equation}
	\mathring{L} = \text{CE}\left(\mathring{y}', y\right)
\end{equation}
where $\text{CE}(\cdot)$ denotes the cross-entropy loss. Subsequently, incomplete modalities are fed into the network for classification:
\begin{equation}
	\widetilde{L}_{\text{task}} = \text{CE}\left(\widetilde{y}', y\right)
\end{equation}

To enable multi-layer self-distillation, we jointly optimize $\widetilde{L}_{\text{task}}$ alongside three distillation losses—$L_{\text{recon}}$, $L_{\text{attn}}$, and $L_{\text{joint}}$—resulting in the overall optimization objective for incomplete modality inputs:
\begin{equation}
	\widetilde{L} = \widetilde{L}_{\text{task}} + \alpha L_{\text{recon}} + \beta L_{\text{attn}} + \gamma L_{\text{joint}}
\end{equation}
where $\alpha$, $\beta$, and $\gamma$ are hyperparameters that control the contribution of each distillation component.

\section{Experimental Setup}
\subsection{Datasets}
\begin{table}[!t]
	\caption{Statistics of datasets.}
	\label{dataset}
	\centering
	\tiny
	\begin{tabular}{ccccccc}
		\toprule
		\multicolumn{4}{c}{Dataset}                                                                                & Training               & Validation            & Test       \\ \midrule
		\multirow{5}{*}{CMU-MOSI}  & \multicolumn{3}{c}{IID~\cite{zadeh2016multimodal}}                                                       & 1284                   & 229                   & 686        \\ \cmidrule{2-7} 
		& \multirow{4}{*}{OOD} & \multirow{2}{*}{binary}  & Sun et al.~\cite{sun2022counterfactual}                  & 1528                   & 270                   & 400        \\ \cmidrule{4-7} 
		&                      &                          & Ours                        & 1284                   & 229                   & 686        \\ \cmidrule{3-7} 
		&                      & \multirow{2}{*}{7-class} & Sun et al.~\cite{sun2022counterfactual}                  & 1528                   & 270                   & 400        \\ \cmidrule{4-7} 
		&                      &                          & Ours                        & 1284                   & 229                   & 686        \\ \midrule
		\multirow{7}{*}{CMU-MOSEI} & \multicolumn{3}{c}{IID~\cite{zadeh2018multimodal}}                                                       & 16326                  & 1871                  & 4659       \\ \cmidrule{2-7} 
		& \multirow{6}{*}{OOD} & \multirow{3}{*}{binary}  & \multirow{2}{*}{Sun et al.~\cite{sun2022counterfactual}} & \multirow{2}{*}{16957} & \multirow{2}{*}{1848} & 2467 (IID) \\ \cmidrule{7-7} 
		&                      &                          &                             &                        &                       & 1715 (OOD) \\ \cmidrule{4-7} 
		&                      &                          & Ours                        & 16326                  & 1871                  & 4659       \\ \cmidrule{3-7} 
		&                      & \multirow{3}{*}{7-class} & \multirow{2}{*}{Sun et al.~\cite{sun2022counterfactual}} & \multirow{2}{*}{16770} & \multirow{2}{*}{1833} & 2282 (IID) \\ \cmidrule{7-7} 
		&                      &                          &                             &                        &                       & 1955 (OOD) \\ \cmidrule{4-7} 
		&                      &                          & Ours                        & 16326                  & 1872                  & 4658       \\ \bottomrule
	\end{tabular}
\end{table}
\begin{figure}[!t]
	\centering
	\includegraphics[width=\linewidth]{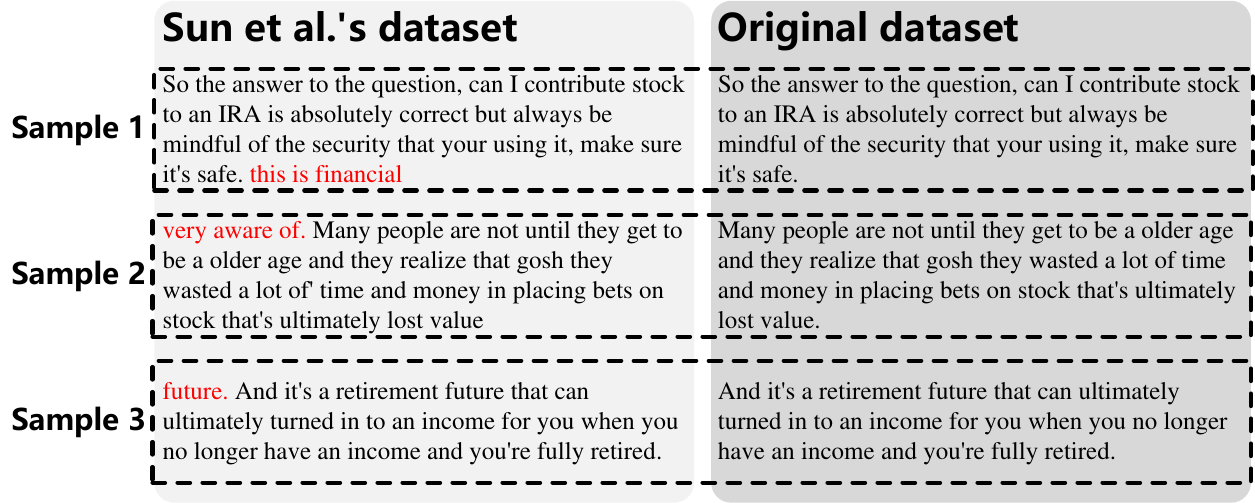}
	\caption{The phenomenon of mixed sentences in the MER OOD datasets partitioned by Sun et al.~\cite{sun2022counterfactual}. The sentences highlighted in red indicate that they come from the beginning of the next sample or the end of the previous sample.}
	\label{dataset_des}
\end{figure}
\textbf{CMU-MOSI~\cite{zadeh2016multimodal}:} A dataset consisting of 2,199 YouTube movie review videos annotated with sentiment scores ranging from -3 (strongly negative) to +3 (strongly positive). The dataset includes 93 speakers and is divided into 1,284 training, 229 validation, and 686 test videos.
\textbf{CMU-MOSEI~\cite{zadeh2018multimodal}:} A larger dataset comprising 22,777 videos, similarly annotated with sentiment scores and split into 16,265 training, 1,869 validation, and 4,643 test videos.

To construct the OOD datasets with unaligned multimodal sequences, we repartitioned the original CMU-MOSI and CMU-MOSEI datasets. The statistics for the original MER IID dataset, the MER OOD dataset partitioned by Sun et al.~\cite{sun2022counterfactual}, and our repartitioned MER OOD dataset are presented in Table~\ref{dataset}.

Additionally, we provide three examples to illustrate the phenomenon of mixed sentences in the MER OOD dataset partitioned by Sun et al.~\cite{sun2022counterfactual}, as depicted in Fig.~\ref{dataset_des}.

\subsection{Feature Extraction}
\textbf{Language:} Text features are extracted using a fine-tuned BERT model~\cite{devlin2019bert}, with a feature dimension of 768. The sequences are padded to a maximum length of 50 tokens.
\textbf{Audio:} Audio features are extracted using COVAREP~\cite{degottex2014covarep}, with dimensions of 5 for CMU-MOSI and 74 for CMU-MOSEI. The sequences are padded to lengths of 375 for CMU-MOSI and 500 for CMU-MOSEI.
\textbf{Visual:} Visual features are extracted using Facet~\footnote{iMotions 2017. https://imotions.com/}, with dimensions of 20 for CMU-MOSI and 35 for CMU-MOSEI. The sequences are padded to a uniform length of 500 for both datasets.

\subsection{Evaluation Metrics}
We evaluate performance using three metrics: Acc2 (binary sentiment classification), F1 score, and Acc7 (7-class sentiment classification). For incomplete data, we calculate the Area Under Indicators Line Chart (AUILC)~\cite{yuan2021transformer} to assess overall performance across increasing missing rates. The AUILC is computed as follows:
\begin{equation}
	v^\diamondsuit = \sum_{i=0}^{t} \frac{1}{2} \left(v_i + v_{i+1}\right)\left(p_{i+1} - p_i\right)
\end{equation}  
where $\{v_0, v_1, \ldots, v_t\}$ ($v \in \{\text{Acc2}, \text{F1}, \text{Acc7}\}$) represents the model's performance under different missing rates $\{p_0, p_1, \ldots, p_t\}$. To maintain consistency with prior works~\cite{yuan2021transformer, sun2023efficient}, all calculations involving AUILC in this paper are performed over a missing rate range of 0.0 to 1.0.

\subsection{Implementation Details}
\begin{table}[!t]
	\caption{Hyperparameters fixed and tuned by Optuna.}
	\label{params}
	\centering
	\scriptsize
	\begin{tabular}{cccc}
		\toprule
		\multicolumn{4}{c}{Fixed Hyperparameters}                                \\ \midrule
		Batch Size                 & \multicolumn{3}{c}{128}                     \\ 
		Epochs                     & \multicolumn{3}{c}{200}                     \\ 
		Early Stop                 & \multicolumn{3}{c}{10}                      \\ 
		Initial Learning Rate      & \multicolumn{3}{c}{1e-3}                    \\ 
		Hidden Unit Size $d$           & \multicolumn{3}{c}{32}                      \\ \midrule
		\multicolumn{4}{c}{Hyperparameters Tuned by Optuna}                      \\ \midrule
		& Range            & Step Size & Distribution \\ \midrule
		MCT Layers                 & {[}1, 4{]}       & 1         & -            \\ 
		Attention Heads            & {[}1, 2, 4, 8{]} & -         & Discrete     \\ 
		Attention Dropout          & {[}0.0, 0.5{]}   & 0.1       & -            \\ 
		Language Embedding Dropout & {[}0.0, 0.5{]}   & 0.1       & -            \\ 
		Output Dropout             & {[}0.0, 0.5{]}   & 0.1       & -            \\ 
		$L_{\text{recon}}$ Weight $\alpha$          & {[}0.0, 1.0{]}   & 0.1       & -            \\ 
		$L_{\text{attn}}$ Weight $\beta$          & {[}0.0, 1.0{]}   & 0.1       & -            \\ 
		$L_{\text{joint}}$ Weight $\gamma$           & {[}0.0, 1.0{]}   & 0.1       & -            \\ \bottomrule
	\end{tabular}
\end{table}
The experiments were conducted on an NVIDIA Quadro RTX 8000 GPU, using Python 3.10, PyTorch 2.5.1, and Adam~\cite{kingma2014adam} as the optimizer. Hyperparameter optimization was performed using Optuna~\cite{akiba2019optuna}. The fixed hyperparameters and those requiring tuning via Optuna are detailed in Table~\ref{params}.

\subsection{Baselines}
\begin{table}[!t]
	\caption{Statistics of the data types handled by baselines and CIDer. ``One-to-one": the model needs to be trained and tested at each missing rate. ``One-to-all": the model is trained only once and then tested at each missing rate.}
	\label{statistic}
	\centering
	\tiny
	\begin{tabular}{ccccccc}
		\toprule
		\multirow{3}{*}{Models} & \multirow{3}{*}{\begin{tabular}[c]{@{}c@{}}Complete\\ Data\end{tabular}} & \multicolumn{3}{c}{Incomplete Data}                                & \multirow{3}{*}{\begin{tabular}[c]{@{}c@{}}OOD\\ Data\end{tabular}} & \multirow{3}{*}{\begin{tabular}[c]{@{}c@{}}Unaligned\\ Data\end{tabular}} \\ \cmidrule{3-5}
		&                                                                             & Traditional               & \multirow{2}{*}{RMM}      & Training   &                                                                     &                                                                           \\
		&                                                                             & RMFM                      &                           & Strategy   &                                                                     &                                                                           \\ \midrule
		DMD~\cite{li2023decoupled}                     & \Checkmark                                                   &                           &                           &            &                                                                     & \Checkmark                                                 \\
		GLoMo~\cite{zhuang2024glomo}                   & \Checkmark                                                   &                           &                           &            &                                                                     & \Checkmark                                                 \\
		DLF~\cite{wang2024dlf}                     & \Checkmark                                                   &                           &                           &            &                                                                     & \Checkmark                                                 \\ \midrule
		EMT-DLFR~\cite{sun2023efficient}                & \Checkmark                                                   & \Checkmark &                           & one-to-one &                                                                     & \Checkmark                                                 \\
		LNLN~\cite{zhang2024towards}                    & \Checkmark                                                   & \Checkmark &                           & one-to-all &                                                                     & \Checkmark                                                 \\ \midrule
		DiCMoR~\cite{wang2023distribution}                  & \Checkmark                                                   &                           & \Checkmark & one-to-one &                                                                     &                                                                           \\
		MPLMM~\cite{guo2024multimodal}                   & \Checkmark                                                   &                           & \Checkmark & one-to-all &                                                                     & \Checkmark                                                 \\ \midrule
		CLUE~\cite{sun2022counterfactual}                    & \Checkmark                                                   &                           &                           &            & \Checkmark                                           &                                                                           \\
		GEAR~\cite{sun2023general}                    & \Checkmark                                                   &                           &                           &            & \Checkmark                                           & \Checkmark                                                 \\ \midrule
		\rowcolor{gray!20}
		CIDer (Ours)            & \Checkmark                                                   & \Checkmark & \Checkmark & one-to-all & \Checkmark                                           & \Checkmark                                                 \\ \bottomrule
	\end{tabular}
\end{table}
We compare our proposed CIDer with state-of-the-art methods, categorized into four groups based on the data types they handle:
1) Full modality input: \textbf{DMD}~\cite{li2023decoupled}, \textbf{GLoMo}~\cite{zhuang2024glomo}, \textbf{DLF}~\cite{wang2024dlf}.
2) Traditional RMFM: \textbf{EMT-DLFR}~\cite{sun2023efficient}, \textbf{LNLN}~\cite{zhang2024towards}.
3) RMM: \textbf{DiCMoR}~\cite{wang2023distribution}, \textbf{MPLMM}~\cite{guo2024multimodal}.
4) OOD data: \textbf{CLUE}~\cite{sun2022counterfactual} (using MAG-BERT~\cite{rahman2020integrating} as the base model), \textbf{GEAR}~\cite{sun2023general}.

For DiCMoR and MAG-BERT, multimodal sequences were aligned using Connectionist Temporal Classification (CTC)~\cite{graves2006connectionist} prior to input. Table~\ref{statistic} summarizes the data types handled by each baseline and CIDer.

\section{Results and Discussion}
\subsection{Comparison to State-of-the-art}
\begin{table*}[!t]
	\caption{Comparison with state-of-the-art methods. ``RMFM" refers to incomplete data input where the missing modality type corresponds to Random Modality Feature Missing, while ``-" indicates complete data input. ``UA" denotes unaligned multimodal sequences in the dataset. ``IID" signifies that the test set distribution matches that of the training and validation sets, whereas ``OOD" indicates that the test set distribution differs from that of the training and validation sets.}
	\label{sota}
	\centering
	\tiny
	\begin{tabular}{ccccccccccccccccccc}
		\toprule
		\multirow{3}{*}{Models} & \multicolumn{3}{c}{CMU-MOSI}      & \multicolumn{3}{c}{CMU-MOSEI}     & \multicolumn{3}{c}{CMU-MOSI}   & \multicolumn{3}{c}{CMU-MOSEI}  & \multicolumn{3}{c}{CMU-MOSI}      & \multicolumn{3}{c}{CMU-MOSEI}     \\
		& \multicolumn{3}{c}{(RMFM/UA/IID)} & \multicolumn{3}{c}{(RMFM/UA/IID)} & \multicolumn{3}{c}{(-/UA/OOD)} & \multicolumn{3}{c}{(-/UA/OOD)} & \multicolumn{3}{c}{(RMFM/UA/OOD)} & \multicolumn{3}{c}{(RMFM/UA/OOD)} \\ \cmidrule{2-19} 
		& Acc2$^\diamondsuit$       & F1$^\diamondsuit$       & Acc7$^\diamondsuit$      & Acc2$^\diamondsuit$        & F1$^\diamondsuit$      & Acc7$^\diamondsuit$      & Acc2      & F1      & Acc7     & Acc2       & F1     & Acc7     & Acc2$^\diamondsuit$       & F1$^\diamondsuit$       & Acc7$^\diamondsuit$      & Acc2$^\diamondsuit$        & F1$^\diamondsuit$      & Acc7$^\diamondsuit$      \\ \midrule
		DMD~\cite{li2023decoupled}                     & 63.7           & 57.8         & 29.4          & 76.1            & 74.8        & 45.9          & 74.0          & 74.1        & 42.7         & 73.0           & 73.2       & 49.0         & 60.3           & 60.2         & 28.1          & 59.3            & 57.7        & 42.2          \\
		GLoMo~\cite{zhuang2024glomo}                   & 67.6           & 65.0         & 27.7          & 76.5            & 75.1        & 44.5          & 76.0          & 76.1        & 43.6         & 73.8           & 74.0       & 49.3         & 60.5           & 60.4         & 27.2          & 59.7            & 58.4        & 42.2          \\
		DLF~\cite{wang2024dlf}                     & 66.4
		& 63.7         & 28.7          & 76.0            & 74.2        & 45.8          & 72.2          & 72.0        & 42.7         & 74.0           & 74.3       & 48.8         & 59.9           & 59.9         & 27.7          & 59.2            & 57.2        & 42.6          \\ \midrule
		EMT-DLFR~\cite{sun2023efficient}                & 70.4           & 70.1         & 33.0          & 74.4            & 73.6        & 45.7          & 74.8          & 74.9        & 43.4         & 71.2           & 71.3       & \textbf{49.8}         & 62.3           & 62.4         & 29.9          & 60.1            & 59.2        & 42.0          \\
		LNLN~\cite{zhang2024towards}                    & 70.4           & 70.5         & 30.8          & \textbf{77.0}            & \textbf{79.0}        & 44.5          & 77.1          & 77.0        & 43.0         & 76.8           & 76.8       & 49.2         & 62.4           & 62.3         & 29.8          & 64.3            & 64.2        & 40.9          \\ \midrule
		DiCMoR~\cite{wang2023distribution}                  & 69.8           & 69.4         & 30.4          & 74.9            & 73.5        & 46.4          & 74.5          & 74.5        & 41.0         & 73.8           & 74.0       & 49.0         & 61.2           & 60.7         & 29.6          & 60.4            & 59.2        & 43.2          \\
		MPLMM~\cite{guo2024multimodal}                   & 66.3           & 66.2         & 24.9          & 73.6            & 74.8        & 43.0          & 70.9          & 70.9        & 25.1         & 68.4           & 68.1       & 38.2         & 60.4           & 60.3         & 22.6          & 57.6            & 59.9        & 40.0          \\ \midrule
		CLUE~\cite{sun2022counterfactual}                    & 67.3           & 67.8         & 28.3          & 76.1            & 77.9        & 46.8          & 73.7          & 73.6        & 39.7         & 73.0           & 72.9       & 48.7         & 59.4           & 59.4         & 28.5          & 61.3            & 61.3        & 42.7          \\
		GEAR~\cite{sun2023general}                    & 62.6           & 58.2         & 26.2          & 71.0            & 63.2        & 44.7          & 71.6          & 71.2        & 42.7         & 69.8           & 69.7       & 47.7         & 59.1           & 58.2         & 26.8          & 54.3            & 48.9        & 41.2          \\ \midrule
		DLF+MACI                & 69.4           & 68.5         & 30.4          & 76.0            & 74.8        & 45.3          & 74.6          & 74.7        & 40.7         & 76.2           & 76.4       & 48.5         & 62.4           & 61.2         & 29.3          & 64.4            & 63.7        & 42.3          \\
		MPLMM+MACI              & 67.9           & 67.1         & 27.4          & 70.4            & 71.2        & 43.6          & 74.6          & 74.7        & 35.3         & 73.9           & 73.6       & 42.2         & 62.2           & 61.6         & 26.8          & 63.0            & 62.6        & 40.3          \\ \midrule
		\rowcolor{gray!20}
		CIDer (Ours)            & \textbf{72.1}           & \textbf{71.4}         & \textbf{33.6}          & 76.0            & 75.4        & \textbf{48.0}          & \textbf{78.1}          & \textbf{78.2}        & \textbf{45.0}         & \textbf{78.7}           & \textbf{79.5}       & 49.6         & \textbf{65.8}           & \textbf{63.9}         & \textbf{32.0}          & \textbf{64.6}
		& \textbf{65.8}
		& \textbf{43.6}          \\ \bottomrule
	\end{tabular}
\end{table*}
\textbf{The RMFM scenario:} To evaluate CIDer's performance under modality-missing scenarios, we conducted experiments on the unaligned CMU-MOSI and CMU-MOSEI datasets with RMFM as the missing type. As shown on the left side of Table~\ref{sota}, among the six metrics across the two datasets, CIDer achieved the best performance in four of them. It fell behind LNLN~\cite{zhang2024towards} in terms of Acc2 and F1 scores on CMU-MOSEI. This is likely because LNLN has a significantly larger number of model parameters compared to CIDer, enabling it to exhibit more robust performance across varying missing rates.

\textbf{The OOD Scenario:} To assess CIDer's performance in OO) scenarios, we conducted experiments on the same two datasets, as shown in the middle section of Table~\ref{sota}. The results demonstrate that under the OOD condition with complete modality input, CIDer achieved the best performance on all metrics for both datasets, except for Acc7 on CMU-MOSEI, where it was only 0.2\% lower than the optimal result obtained by EMT-DLFR~\cite{sun2023efficient}.

\textbf{The RMFM and OOD scenario:} To further validate CIDer's robustness when addressing both RMFM and OOD challenges simultaneously, we conducted experiments on the above datasets with RMFM as the modality-missing type. As shown on the right side of Table~\ref{sota}, the results indicate that CIDer effectively handles both RMFM and OOD challenges, outperforming various state-of-the-art methods on both datasets.

To validate the effectiveness of the MACI module when integrated into different MER methods, we conducted experiments on DLF~\cite{wang2024dlf} and MPLMM~\cite{guo2024multimodal}. As demonstrated in Table~\ref{sota}, incorporating MACI improved the performance of both DLF and MPLMM across most scenarios. These results further confirm the plug-and-play portability and efficacy of MACI.

Extensive experiments demonstrate that CIDer, by integrating self-distillation and causal inference, maintains high robustness even in the presence of both RMFM and OOD challenges. Although CIDer does not outperform LNLN in some scenarios, its parameter count is significantly lower, comprising approximately only 3.8\% of LNLN's parameters, as detailed in Subsection~\ref{complexity analysis}. This makes CIDer a highly efficient and practical solution for real-world MER applications.

\subsection{Ablation Study}
\begin{table}[!t]
	\caption{Ablation studies. ``F.C.T." refers to Fine-grained Counterfactual Text.}
	\label{ablation}
	\centering
	\footnotesize
	\begin{tabular}{ccccc}
		\toprule
		\multicolumn{2}{c}{\multirow{3}{*}{Models}}       & \multicolumn{3}{c}{CMU-MOSI}      \\
		\multicolumn{2}{c}{}                              & \multicolumn{3}{c}{(RMFM/UA/OOD)} \\ \cmidrule{3-5} 
		\multicolumn{2}{c}{}                              & Acc2$^\diamondsuit$       & F1$^\diamondsuit$       & Acc7$^\diamondsuit$      \\ \midrule
		\rowcolor{gray!20}
		\multicolumn{2}{c}{CIDer}                         & \textbf{65.8}           & \textbf{63.9}         & \textbf{32.0}          \\ \midrule
		\multirow{4}{*}{w/o MSSD} & w/o $L_{\text{recon}}$ & 63.4           & 62.4         & 31.8          \\
		& w/o $L_{\text{attn}}$     & 64.3           & 63.3         & 30.3          \\
		& w/o $L_{\text{joint}}$     & 64.4           & 61.8         & 31.7          \\
		& w/o all       & 62.5           & 61.5         & 31.1          \\ \midrule
		\multirow{3}{*}{w/o MACI}       & w/o MCM      & 61.7           & 61.3         & 31.0          \\
		& w/o F.C.T.      & 62.1           & 60.9         & 30.8          \\
		& w/o all       & 64.1           & 62.1         & 31.2          \\ \midrule
		\multicolumn{2}{c}{w/o WSAM}                                         & 63.4           & 62.6         & 29.6          \\
		\multicolumn{2}{c}{w/o MCT}                                          & 64.3           & 62.7         & 31.4          \\ \bottomrule
	\end{tabular}
\end{table}
To evaluate the performance of each component in the proposed CIDer framework, we conducted experiments on the CMU-MOSI dataset, as shown in Table~\ref{ablation}. The results indicate that both self-distillation and causal inference are crucial for addressing RMFM and OOD challenges. The absence of either module leads to a noticeable decline in performance.

Furthermore, the handling of lengthy non-linguistic sequences and the fusion of multimodal information also have a significant impact on the model's performance.
Row 9 (w/o WSAM) illustrates that even with the full architecture, the model's performance is still compromised when processing lengthy non-linguistic sequences. This suggests that such sequences often contain redundant information, which can adversely affect the model's judgment.
Row 10 (w/o MCT) demonstrates that replacing MCT with a vanilla transformer results in performance degradation. This highlights that MCT improves the model’s efficiency by constructing distinct attention matrices, enabling effective fusion of multimodal information in a simple yet powerful manner.

\subsection{Complexity Analysis}
\label{complexity analysis}
\begin{table}[!t]
	\caption{Complexity Analysis. *: To make the results more significant, we subtracted the parameters of BERT$_\text{BASE}$.}
	\label{complexity}
	\centering
	\scriptsize
	\begin{tabular}{cccc}
		\toprule
		\multirow{2}{*}{Models} & \multirow{2}{*}{Parameters*} & GPU Memory Usage & Training Time \\
		&                             & (GiB)      & (s/epoch)     \\ \midrule
		DMD~\cite{li2023decoupled}                     & 12,578,232                            & 29.3           & 72.1              \\
		GLoMo~\cite{zhuang2024glomo}                   & 322,823                            & 26.7           & 10.4              \\
		DLF~\cite{wang2024dlf}                     & 10,750,808                            & 19.9           & 56.1              \\ \midrule
		EMT-DLFR~\cite{sun2023efficient}                & 2,322,779                            & 11.8           & 11.8              \\
		LNLN~\cite{zhang2024towards}                    & 6,483,012                            & 13.2           & 12.6              \\ \midrule
		DiCMoR~\cite{wang2023distribution}                  & 3,534,129                            & \textbf{1.8}           & 27.8              \\
		MPLMM~\cite{guo2024multimodal}                   & 2,242,946                            & 15.6           & 31.2              \\ \midrule
		CLUE~\cite{sun2022counterfactual}                    & 3,798,579                            & 2.6           & 22.8              \\
		GEAR~\cite{sun2023general}                    & 372,071                            & 7.9           & \textbf{8.8}              \\ \midrule
		MACI                    & 37,405                            & -           & -              \\
		\rowcolor{gray!20}
		CIDer (Ours)            & \textbf{247,946}                            & 6.8           & 10.4              \\ \bottomrule
	\end{tabular}
\end{table}
To further evaluate the spatial and temporal complexity of the proposed CIDer, we computed the model parameters, GPU memory usage, and training time for both baseline methods and CIDer on the CMU-MOSI dataset, as presented in Table~\ref{complexity}. The results indicate that CIDer has the fewest parameters among all state-of-the-art methods, with approximately 248K parameters. Furthermore, CIDer's GPU memory usage ranks second only to DiCMoR~\cite{wang2023distribution} and CLUE~\cite{sun2022counterfactual} optimized with CTC, consuming only 6.8 GiB. We attribute the significantly lower memory usage of DiCMoR and CLUE to the fact that CTC performs word-level alignment of non-linguistic sequences outside the network at a relatively low computational cost. However, CTC-based training remains unstable, with the CTC loss frequently reaching magnitudes on the order of $10^4$.

Additionally, CIDer demonstrates a relatively short training time, surpassed only by GEAR~\cite{sun2023general}, which requires fine-tuning two BERT$_\text{BASE}$ models, and is comparable to GLoMo~\cite{zhuang2024glomo}, averaging 10.4 seconds per epoch.

We also validated the number of parameters in the introduced debiasing module MACI, which amounts to approximately 37K. This indicates that MACI can be integrated as an auxiliary module into existing MER methods with minimal additional parameters while enhancing their generalization capabilities in OOD environments.

In summary, despite incorporating multiple modules, CIDer remains a compact model characterized by a low parameter count, efficient memory usage, and fast training speed.

\subsection{Cross-Dataset Generalization}
\begin{table}[!t]
	\caption{Cross-dataset generalization. ``A" denotes that the mutlimodal sequences in the dataset are aligned.}
	\label{cross_dataset}
	\centering
	\scriptsize
	\begin{tabular}{ccccccc}
		\toprule
		\multirow{4}{*}{Models} & \multicolumn{3}{c}{CMU-MOSI}                & \multicolumn{3}{c}{CMU-MOSEI}              \\
		& \multicolumn{3}{c}{$\rightarrow$ CMU-MOSEI} & \multicolumn{3}{c}{$\rightarrow$ CMU-MOSI} \\
		& \multicolumn{3}{c}{(RMFM/A/IID)}           & \multicolumn{3}{c}{(RMFM/A/IID)}          \\ \cmidrule{2-7} 
		& Acc2$^\diamondsuit$          & F1$^\diamondsuit$          & Acc7$^\diamondsuit$          & Acc2$^\diamondsuit$           & F1$^\diamondsuit$         & Acc7$^\diamondsuit$         \\ \midrule
		GLoMo~\cite{zhuang2024glomo}                   & 68.7              & 66.3            & 38.3              & 67.5               & 66.8           & 27.7             \\ \midrule
		EMT-DLFR~\cite{sun2023efficient}                & 62.1              & 61.3            & 33.4              & 64.9               & 63.8           & 26.1             \\
		LNLN~\cite{zhang2024towards}                    & \textbf{71.4}              & \textbf{73.6}            & 32.5              & 67.5               & 67.6           & 29.5             \\ \midrule
		DiCMoR~\cite{wang2023distribution}                  & 69.3              & 68.1            & 31.3              & 64.1               & 62.6           & 26.3             \\ \midrule
		CLUE~\cite{sun2022counterfactual}                    & 65.9              & 66.2            & 28.6              & 67.4               & 68.4           & 29.4             \\
		GEAR~\cite{sun2023general}                    & 69.1              & 63.6            & 36.5              & 58.0               & 51.4           & 24.0             \\ \midrule
		\rowcolor{gray!20}
		CIDer (Ours)            & 70.0              & 66.3            & \textbf{42.4}              & \textbf{69.6}               & \textbf{68.5}           & \textbf{32.5}             \\ \bottomrule
	\end{tabular}
\end{table}
To investigate the generalization capability of CIDer in cross-dataset scenarios (domain generalization, which is a generalized form of OOD), we conducted experiments on the CMU-MOSI and CMU-MOSEI datasets. These datasets were re-partitioned by Guo et al.~\cite{guo2024bridging} to ensure compatibility in feature dimensions. Specifically, CMU-MOSI $\rightarrow$ CMU-MOSEI denotes training on the CMU-MOSI training set and testing on the CMU-MOSEI test set, while CMU-MOSEI $\rightarrow$ CMU-MOSI follows the same reasoning.

As shown in Table~\ref{cross_dataset}, CIDer achieved the highest performance in four out of six metrics across the two datasets. It fell only in terms of Acc2 and F1 scores for the CMU-MOSI $\rightarrow$ CMU-MOSEI scenario, where LNLN~\cite{zhang2024towards} outperformed it. We attribute this to LNLN's larger parameter count, which may enhance its generalization in scenarios involving missing modalities. 

Overall, despite not being explicitly designed for domain generalization, CIDer demonstrated superior performance in most cases while maintaining a lower parameter count. This further underscores its robustness and efficiency.

\subsection{Other Modality Missing Scenarios}
\begin{table*}[!t]
	\caption{Comparison with state-of-the-art methods. ``Traditional RMFM" refers to incomplete data input where the missing modality type corresponds to Traditional Random Modality Feature Missing. ``RMM" refers to incomplete data input where the missing modality type corresponds to Random Modality Missing. ``TMFM" refers to incomplete data input where the missing modality type corresponds to Temporal Modality Feature Missing. ``STMFM" refers to incomplete data input where the missing modality type corresponds to Structural Temporal Modality Feature Missing.}
	\label{other_missing_scenarios}
	\centering
	\begin{tabular}{ccccccccccccc}
		\toprule
		\multirow{3}{*}{Models} & \multicolumn{3}{c}{CMU-MOSI} & \multicolumn{3}{c}{CMU-MOSI} & \multicolumn{3}{c}{CMU-MOSI}     & \multicolumn{3}{c}{CMU-MOSI}      \\
		& \multicolumn{3}{c}{(Traditional RMFM/UA/IID)} & \multicolumn{3}{c}{(RMM/UA/IID)} & \multicolumn{3}{c}{(TMFM/A/IID)} & \multicolumn{3}{c}{(STMFM/A/IID)} \\ \cmidrule{2-13} 
		& Acc2$^\diamondsuit$       & F1$^\diamondsuit$       & Acc7$^\diamondsuit$ & Acc2$^\diamondsuit$       & F1$^\diamondsuit$       & Acc7$^\diamondsuit$ & Acc2$^\diamondsuit$       & F1$^\diamondsuit$       & Acc7$^\diamondsuit$     & Acc2$^\diamondsuit$       & F1$^\diamondsuit$        & Acc7$^\diamondsuit$     \\ \midrule
		DMD~\cite{li2023decoupled}                     & 65.3 & 61.1 & 29.3 & 66.2 & 63.6 & 29.5 & 67.6       & 65.1     & 31.1     & 67.0       & 64.3      & 29.6     \\
		GLoMo~\cite{zhuang2024glomo}                   & 68.2 & 67.2 & 29.3 & 65.2 & 63.4 & 28.6 & 66.3       & 63.6     & 28.1     & 67.7       & 65.0      & 29.8     \\
		DLF~\cite{wang2024dlf}                     & 68.0 & 66.1 & 29.6 & 65.7 & 64.0 & 29.7 & 66.4       & 65.0     & 28.4     & 65.9       & 61.7      & 28.3     \\ \midrule
		EMT-DLFR~\cite{sun2023efficient}                & 70.3 & 70.4 & 32.4 & 69.6 & 69.6 & 30.9 & 70.8       & 70.7     & 33.0     & 71.3       & 71.3      & 32.6     \\
		LNLN~\cite{zhang2024towards}                    & 70.5 & 70.6 & 30.4 & \textbf{71.5} & \textbf{72.5} & 31.9 & 70.2       & 70.4     & 30.5     & 69.0       & 69.3      & 30.4     \\ \midrule
		DiCMoR~\cite{wang2023distribution}                  & 69.1 & 68.9 & 28.6 & 66.7 & 65.0 & 29.9 & 66.6       & 65.7     & 28.5     & 66.2       & 63.5      & 29.4     \\
		MPLMM~\cite{guo2024multimodal}                   & 63.4 & 63.8 & 19.4 & 60.4 & 64.2 & 23.8 & 61.7       & 64.0     & 22.3     & 63.7       & 65.3      & 25.7     \\ \midrule
		CLUE~\cite{sun2022counterfactual}                    & 65.9 & 66.3 & 29.2 & 65.2 & 65.9 & 28.5 & 66.3       & 66.9     & 29.2     & 66.9       & 68.2      & 30.1     \\
		GEAR~\cite{sun2023general}                    & 64.9 & 60.7 & 27.2 & 62.9 & 59.6 & 30.3 & 64.7       & 60.4     & 27.3     & 65.3       & 61.0      & 28.5     \\ \midrule
		\rowcolor{gray!20}
		CIDer (Ours)            & \textbf{71.7} & \textbf{71.0} & \textbf{33.3} & 70.5 & 68.5 & \textbf{33.2} & \textbf{72.0}       & \textbf{71.9}     & \textbf{33.5}     & \textbf{72.3}       & \textbf{72.2}      & \textbf{32.7}     \\ \bottomrule
	\end{tabular}
\end{table*}
\begin{table*}[!t]
	\caption{Comparison with state-of-the-art methods. ``SMM" refers to incomplete data input where the missing modality type corresponds to Specific Modality Missing.}
	\label{smm_ua_iid}
	\centering
	\tiny
	\begin{tabular}{ccccccccccccccccccc}
		\toprule
		\multirow{4}{*}{Models} & \multicolumn{18}{c}{CMU-MOSI}                                                   \\
		& \multicolumn{18}{c}{(SMM/UA/IID)}                                               \\ \cmidrule{2-19} 
		& \multicolumn{3}{c}{L}    & \multicolumn{3}{c}{A}    & \multicolumn{3}{c}{V} & \multicolumn{3}{c}{L, A} & \multicolumn{3}{c}{L, V} & \multicolumn{3}{c}{A, V}   \\ \cmidrule{2-19} 
		& Acc2   & F1     & Acc7   & Acc2   & F1     & Acc7   & Acc2   & F1     & Acc7 & Acc2   & F1     & Acc7 & Acc2   & F1     & Acc7 & Acc2   & F1     & Acc7   \\ \midrule
		DMD~\cite{li2023decoupled}                     & 84.5   & 84.5   & 45.2   & 42.2   & 25.1   & 15.5   & 54.3   & 53.4   & 16.5 & 83.8   & 83.9   & 45.0   & 85.2   & 85.2   & 47.2   & 42.2   & 25.1   & 15.5  \\
		GLoMo~\cite{zhuang2024glomo}                   & 84.2   & 84.1   & 46.2   & 42.2   & 25.1   & 15.5   & 42.2   & 25.1   & 15.5 & 84.2   & 84.2   & 45.8   & 84.8   & 84.7   & 45.5   & 44.2   & 31.3   & 15.5   \\
		DLF~\cite{wang2024dlf}                     & 84.3   & 84.4   & 46.1   & 52.7   & 48.3   & 15.3   & 42.7   & 26.8   & 15.0 & 84.0   & 84.0   & \textbf{48.7}   & 83.4   & 83.4   & \textbf{48.0}   & 42.2   & 25.1   & 15.5   \\ \midrule
		EMT-DLFR~\cite{sun2023efficient}                & 85.4   & 85.3   & \textbf{46.9}   & 52.3   & 51.4   & 16.5   & 52.7   & 50.0   & 19.2 & 83.8   & 83.8   & 45.9   & 83.8   & 83.8   & 47.1   & 57.0   & 57.0   & 21.3   \\
		LNLN~\cite{zhang2024towards}                    & 85.2   & 85.3   & 43.7   & 56.6   & \textbf{59.8}   & 20.0   & 56.6   & \textbf{59.8}   & 20.0 & 85.2   & 85.3   & 43.7   & 85.2   & 85.3   & 43.7   & 56.6   & 59.8   & 20.0   \\ \midrule
		DiCMoR~\cite{wang2023distribution}                  & 85.5   & 85.5   & 42.9   & 49.9   & 42.6   & 17.1   & 51.1   & 49.0   & 19.0 & 84.9   & 84.9   & 44.2   & 84.9   & 84.9   & 43.4   & 53.4   & 51.4   & 17.5   \\
		MPLMM~\cite{guo2024multimodal}                   & 65.7   & 67.3   & 24.3   & 55.8   & 57.0   & 16.2   & 54.7   & 55.3   & 16.3 & 66.6   & 67.9   & 24.5   & 65.5   & 67.2   & 24.5   & 55.8   & 57.0   & 16.5   \\ \midrule
		CLUE~\cite{sun2022counterfactual}                    & 82.6   & 82.5   & 41.1   & 49.1   & 57.4   & 16.7   & 44.2   & 58.2   & 20.4 & 82.1   & 82.0   & 38.5   & 80.8   & 80.7   & 40.2   & 54.7   & \textbf{65.3}   & \textbf{22.2}   \\
		GEAR~\cite{sun2023general}                    & 83.2   & 83.3   & 44.5   & 42.2   & 25.1   & 15.5   & 42.2   & 25.1   & 15.5 & 82.9   & 83.0   & 44.0   & 80.8   & 80.9   & 45.2   & 42.2   & 25.1   & 15.5   \\ \midrule
		\rowcolor{gray!20}
		CIDer (Ours)            & \textbf{86.3}   & \textbf{86.3}   & 45.5   & \textbf{57.8}   & 50.6   & \textbf{20.1}   & \textbf{57.8}   & 46.3   & \textbf{20.8} & \textbf{86.4}   & \textbf{86.4}   & 45.5   & \textbf{86.3}   & \textbf{86.3}   & 46.2   & \textbf{57.8}   & 51.0   & 21.6  \\ \bottomrule
	\end{tabular}
\end{table*}
To evaluate the performance of CIDer in handling other types of modality missing scenarios (Traditional RMFM, RMM, Temporal Modality Feature Missing (TMFM), Structural Temporal Modality Feature Missing (STMFM)) at different missing rates, we conducted experiments on the CMU-MOSI dataset, as shown in Table~\ref{other_missing_scenarios}. Additionally, we evaluated the performance of CIDer in the special case of RMM, specifically the scenario where a Specific Modality is Missing (SMM), as shown in Table~\ref{smm_ua_iid}. The detailed results for each method at various missing rates under different modality-missing scenarios, along with the final AUILC scores, are provided in the supplementary material.

As shown in Table~\ref{other_missing_scenarios} and Table~\ref{smm_ua_iid}, CIDer maintains robust performance across most cases, even when confronted with various modality-missing scenarios such as Traditional RMFM, RMM, TMFM, STMFM, and SMM. This further highlights the effectiveness and necessity of the hierarchical self-distillation mechanism in CIDer. More importantly, CIDer employs a one-to-all training strategy, which not only ensures strong robustness but also significantly reduces training costs. This characteristic holds substantial practical significance for real-world MER applications.

\subsection{Qualitative Analysis}
\subsubsection{Embedding Space of Joint Representation}
\begin{figure*}[!t]
	\centering
	\includegraphics[width=\linewidth]{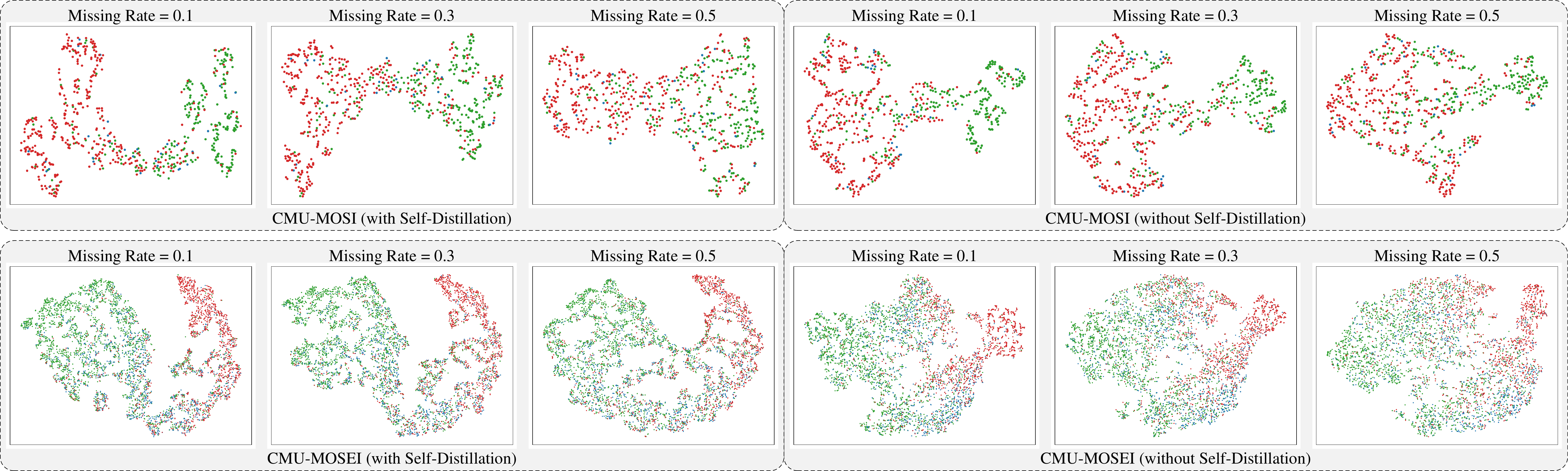}
	\caption{The distribution of joint representations in the embedding space. Red dots correspond to negative samples, green dots represent positive samples, and blue dots indicate neutral samples.}
	\label{joint_rep}
\end{figure*}
To evaluate the effectiveness of self-distillation, we conducted experiments on both the CMU-MOSI and CMU-MOSEI datasets. Scatter plots were generated for missing rates of 0.1, 0.3, and 0.5, as shown in Fig.~\ref{joint_rep}.

From the figures, it is evident that on both the CMU-MOSI and CMU-MOSEI datasets, when self-distillation is applied, the model effectively distinguishes between negative and positive samples despite modality missing. In contrast, when self-distillation is not utilized (as illustrated on the right side of Fig.~\ref{joint_rep}), the separability of features for negative and positive samples is significantly reduced, leading to increased overlap in the embedding space. This overlap adversely affects the final recognition performance.

\subsubsection{Prediction Correction by MACI}
\begin{figure}[!t]
	\centering
	\includegraphics[width=\linewidth]{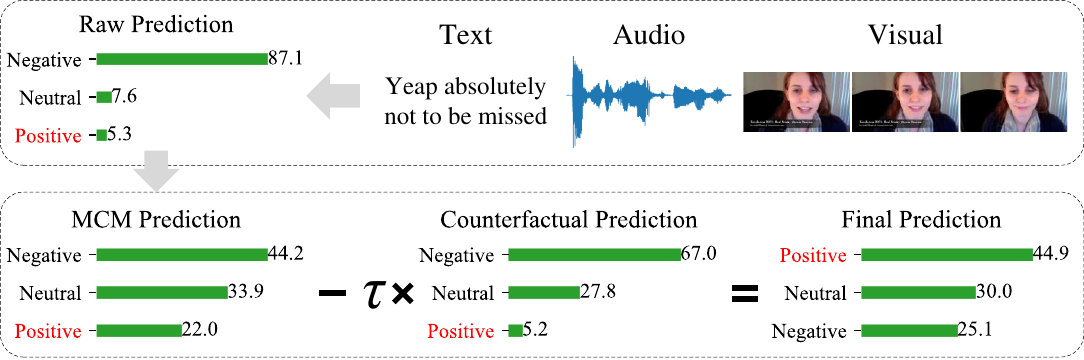}
	\caption{An example of MACI correcting prediction. The category highlighted in red represents the ground truth.}
	\label{maci_prediction}
\end{figure}
To evaluate the effectiveness of the MACI module, we present an example from the CMU-MOSI OOD test set to illustrate its prediction correction process. As shown in Fig.~\ref{maci_prediction}, for a sample labeled as ``positive", the model—relying solely on the statistical information of the dataset—tends to predict it as ``negative" with a high probability of 87.1\%. To enhance the robustness of the model's predictions and mitigate the influence of label bias, the incorporation of the MCM module significantly reduces the model’s prediction probability for the ``negative" class. However, overall, the model still leans toward predicting the emotion expressed by the sample as negative. To further refine the prediction, the model subtracts the results obtained from fine-grained counterfactual text during the test phase, ultimately correctly classifying the sample as positive.

This example demonstrates that, in the context of OOD scenarios, the MACI module effectively alleviates label bias and language bias through the synergy of the MCM and fine-grained counterfactual text. This integration enables accurate predictions and enhances CIDer’s generalization capabilities in OOD scenarios.

\section{Conclusion}
In summary, this paper introduces a novel and robust MER paradigm named CIDer, designed to simultaneously address the challenges of modality missing and OOD data. The paper first defines a more generalized RMFM task and proposes the MSSD module to tackle the RMFM challenge. Specifically, within MSSD, knowledge distillation is performed in a bottom-up manner across three levels: low-level features, attention maps, and high-level representations. To maintain a streamlined overall framework, weight sharing is implemented between the teacher and student models. Additionally, the WSAM module for word-level alignment is introduced to compress lengthy non-linguistic sequences, while the MCT module is incorporated to enable efficient multimodal fusion.

The paper also proposes the MACI module to address OOD challenges. In MACI, a specific causal graph is designed for MER, accounting for the potential impacts of label bias and language bias. The MCM is introduced to mitigate label bias during training. And fine-grained counterfactual texts are constructed to alleviate language bias during testing. Notably, MACI introduces only a small number of learnable parameters and can be independently applied to enhance the OOD generalization capabilities of existing MER models.

{\appendices
	\appendix
	\section*{Definition of Various Modality Missing Scenarios}
	In this section, we mainly provide the mathematical definitions of other modality missing scenarios, which mainly include Traditional Random Modality Feature Missing (Traditional RMFM) and Random Modality Missing (RMM). 
	
	Additionally, according to the research by Yuan et al.~\cite{yuan2023noise}, Traditional RMFM can be further divided into: Temporal Modality Feature Missing (TMFM) and Structural Temporal Modality Feature Missing (STMFM). And RMM can be further divided into Specific Modality Missing (SMM).
	
	\textbf{Traditional RMFM:} For Traditional RMFM, given a missing rate of $r\%$, we can generate missing masks for each modality, $M_L \in \mathbb{R}^{T_l}$, $M_A \in \mathbb{R}^{T_a}$, and $M_V \in \mathbb{R}^{T_v}$, respectively. Thus, $ M_k^{(i)} \in \mathbb{R}^1 = 0$ ($k \in \{L, A, V\}$) indicates that the $i$-th feature in the modality sequence is missing.
	
	\textbf{RMM:} For RMM, given a missing rate of $r\%$, we can generate modality-level missing masks for the three modalities, $M_{\text{all}} = [M_L, M_A, M_V] \in \mathbb{R}^3$, corresponding to the language, audio, and visual modalities, respectively. Thus, $M_k \in \mathbb{R}^1 = 0$ ($k \in \{L, A, V\}$) indicates that all features in the $k$-th modality sequence are completely missing.
	
	\subsection{Special Cases of Traditional RMFM}
	
	\textbf{TMFM (requires aligned multimodal sequences):} For TMFM, given a missing rate of $r\%$, we can generate a missing mask $M \in \mathbb{R}^T$ that is applied to each modality, where $T$ is the unified sequence length after multimodal sequence alignment. Thus, $M^{(i)} \in \mathbb{R}^1 = 0$ indicates that the $i$-th feature is missing in all modality sequences.
	
	\textbf{STMFM (requires aligned multimodal sequences):} STMFM is an extension of TMFM, where information is lost not just at a single time step but over a period. Specifically, for STMFM, given a missing rate of $r\%$, the starting time step for loss is $i$, and the length of the loss period is $t$. Thus, $(M^{(i)}, \dots, M^{(i+t)}) = 0$ indicates that all features from the $i$-th to the ($i + t$)-th are missing in all modality sequences, where $t = T \times r\%$.
	
	\subsection{Special Case of RMM}
	
	\textbf{SMM:} SMM is a special case of RMM, where a specific modality is entirely missing in all samples while the other modalities are retained. For SMM, retaining only the language modality can be described as $M_L = 1$ for all samples, while $M_{\{A, V\}} = 0$. Similarly, retaining only the audio modality can be described as $M_A = 1$ for all samples, while $M_{\{L, V\}} = 0$; retaining only the visual modality can be described as $M_V = 1$ for all samples, while $M_{\{L, A\}} = 0$; retaining only the language and audio modalities can be described as $M_{\{L, A\}} = 1$ for all samples, while $M_V = 0$; retaining only the language and visual modalities can be described as $M_{\{L, V\}} = 1$ for all samples, while $M_A = 0$; retaining only the audio and visual modalities can be described as $M_{\{A, V\}} = 1$ for all samples, while $M_L = 0$.
	
	For detailed definitions of Traditional RMFM, TMFM, and STMFM, please refer to~\cite{yuan2023noise}.
	
	\section*{Experimental Results of the Complete Modality Scenario}
	\begin{table*}[!t]
		\caption{Comparison with state-of-the-art methods.}
		\label{ua_iid}
		\centering
		\begin{tabular}{clcccccc}
			\toprule
			\multicolumn{2}{c}{\multirow{3}{*}{Models}} & \multicolumn{3}{c}{CMU-MOSI}   & \multicolumn{3}{c}{CMU-MOSEI}  \\
			\multicolumn{2}{c}{}                        & \multicolumn{3}{c}{(-/UA/IID)} & \multicolumn{3}{c}{(-/UA/IID)} \\ \cmidrule{3-8} 
			\multicolumn{2}{c}{}                        & Acc2      & F1       & Acc7    & Acc2      & F1       & Acc7    \\ \midrule
			\multicolumn{2}{c}{DMD~\cite{li2023decoupled}}                     & 84.2      & 84.2     & 45.2    & 85.2      & 85.1     & 51.9    \\
			\multicolumn{2}{c}{GLoMo~\cite{zhuang2024glomo}}                   & 84.8      & 84.6     & 46.1    & \textbf{86.2}      & \textbf{86.2}     & 52.6    \\
			\multicolumn{2}{c}{DLF~\cite{wang2024dlf}}                     & 83.8      & 83.8     & \textbf{48.1}    & 83.7      & 83.9     & \textbf{53.8}    \\ \midrule
			\multicolumn{2}{c}{EMT-DLFR~\cite{sun2023efficient}}                & 83.4      & 83.4     & 45.2    & 82.9      & 83.2     & 53.2    \\
			\multicolumn{2}{c}{LNLN~\cite{zhang2024towards}}                    & 85.2      & 85.3     & 43.7    & 85.4      & 85.5     & 52.7    \\ \midrule
			\multicolumn{2}{c}{DiCMoR~\cite{wang2023distribution}}                  & 85.4      & 85.3     & 43.3    & 85.2      & 85.2     & 53.3    \\
			\multirow{3}{*}{MPLMM~\cite{guo2024multimodal}}  & RMFM              & 75.9      & 75.9     & 27.7    & 82.7      & 82.7     & 44.8    \\ \cmidrule{2-8} 
			& Traditional RMFM  & 66.8      & 68.0     & 24.5    & -         & -        & -       \\ \cmidrule{2-8} 
			& RMM               & 79.0      & 78.9     & 30.6    & -         & -        & -       \\ \midrule
			\multicolumn{2}{c}{CLUE~\cite{sun2022counterfactual}}                    & 82.1      & 82.2     & 41.1    & 84.7      & 85.0     & 52.8    \\
			\multicolumn{2}{c}{GEAR~\cite{sun2023general}}                    & 84.8      & 84.7     & 46.8    & 84.7      & 84.5     & 50.8    \\ \midrule
			\multicolumn{2}{c}{DLF+MACI}                & 84.6      & 84.6     & 45.5    & 85.2      & 85.1     & 51.4    \\
			\multicolumn{2}{c}{MPLMM+MACI}              & 81.3      & 81.4     & 32.2    & 82.5      & 82.9     & 47.5    \\ \midrule
			\rowcolor{gray!20}
			\multicolumn{2}{c}{CIDer (Ours)}            & \textbf{86.4}      & \textbf{86.4}     & 46.5    & 85.5      & 85.5     & 53.6    \\ \bottomrule
		\end{tabular}
	\end{table*}
	\begin{table*}[!t]
		\caption{Comparison with state-of-the-art methods.}
		\label{a_iid}
		\centering
		\begin{tabular}{clcccccc}
			\toprule
			\multicolumn{2}{c}{\multirow{3}{*}{Models}} & \multicolumn{3}{c}{CMU-MOSI}  & \multicolumn{3}{c}{CMU-MOSEI} \\
			\multicolumn{2}{c}{}                        & \multicolumn{3}{c}{(-/A/IID)} & \multicolumn{3}{c}{(-/A/IID)} \\ \cmidrule{3-8} 
			\multicolumn{2}{c}{}                        & Acc2      & F1      & Acc7    & Acc2     & F1       & Acc7    \\ \midrule
			\multicolumn{2}{c}{DMD~\cite{li2023decoupled}}                     & 84.0      & 84.0    & 45.9    & 84.8     & 84.8     & 52.7    \\
			\multicolumn{2}{c}{GLoMo~\cite{zhuang2024glomo}}                   & 85.4      & 85.3    & 46.1    & 85.5     & 85.5     & 52.6    \\
			\multicolumn{2}{c}{DLF~\cite{wang2024dlf}}                     & 84.0      & 84.0    & 44.9    & 84.5        & 84.5        & 52.4       \\ \midrule
			\multicolumn{2}{c}{EMT-DLFR~\cite{sun2023efficient}}                & 83.8      & 83.8    & \textbf{47.7}    & 81.9     & 82.2     & 52.2    \\
			\multicolumn{2}{c}{LNLN~\cite{zhang2024towards}}                    & 84.8      & 84.9    & 43.7    & \textbf{85.9}     & \textbf{86.1}     & 52.6    \\ \midrule
			\multicolumn{2}{c}{DiCMoR~\cite{wang2023distribution}}                  & 84.5      & 84.4    & 43.7    & 85.2     & 85.2     & 53.3    \\
			\multirow{3}{*}{MPLMM~\cite{guo2024multimodal}}        & RMFM        & -         & -       & -       & 83.6        & 83.7        & \textbf{59.6}       \\ \cmidrule{2-8} 
			& TMFM        & 72.9      & 72.7    & 28.3    & -        & -        & -       \\ \cmidrule{2-8} 
			& STMFM       & 76.7      & 76.8    & 36.0    & -        & -        & -       \\ \midrule
			\multicolumn{2}{c}{CLUE~\cite{sun2022counterfactual}}                    & 82.7      & 83.0    & 42.4    & 84.4     & 84.9     & 53.2    \\
			\multicolumn{2}{c}{GEAR~\cite{sun2023general}}                    & 84.0      & 83.9    & 45.2    & 82.5     & 81.6     & 48.3    \\ \midrule
			\rowcolor{gray!20}
			\multicolumn{2}{c}{CIDer (Ours)}            & \textbf{86.1}      & \textbf{86.1}    & 46.1    & 85.7     & 85.8     & 53.1    \\ \bottomrule
		\end{tabular}
	\end{table*}
	\begin{table*}[!t]
		\caption{Comparison with state-of-the-art methods.}
		\label{cross_a_iid}
		\centering
		\begin{tabular}{ccccccc}
			\toprule
			\multirow{3}{*}{Models} & \multicolumn{3}{c}{CMU-MOSI $\rightarrow$ CMU-MOSEI} & \multicolumn{3}{c}{CMU-MOSEI $\rightarrow$ CMU-MOSI} \\
			& \multicolumn{3}{c}{(-/A/IID)}            & \multicolumn{3}{c}{(-/A/IID)}            \\ \cmidrule{2-7} 
			& Acc2         & F1          & Acc7        & Acc2         & F1          & Acc7        \\ \midrule
			GLoMo~\cite{zhuang2024glomo}                   & 81.4         & 81.6        & 44.5        & 80.7         & 80.6        & 38.6        \\ \midrule
			EMT-DLFR~\cite{sun2023efficient}                & 78.9         & 79.2        & 43.8        & 81.7         & 81.2        & 38.6        \\
			LNLN~\cite{zhang2024towards}                    & 80.9         & 81.0        & 42.9        & 82.9         & 83.1        & 45.0        \\ \midrule
			DiCMoR~\cite{wang2023distribution}                  & 79.8         & 80.0        & 42.6        & 79.8         & 79.6        & 41.5        \\ \midrule
			CLUE~\cite{sun2022counterfactual}                    & 76.4         & 76.3        & 39.9        & 83.0         & 83.1        & 42.9        \\
			GEAR~\cite{sun2023general}                    & 79.8         & 80.0        & 37.9        & 78.3         & 78.2        & 41.0        \\ \midrule
			\rowcolor{gray!20}
			CIDer (Ours)            & \textbf{82.7}         & \textbf{82.5}        & \textbf{47.4}        & \textbf{84.4}         & \textbf{84.4}        & \textbf{45.4}        \\ \bottomrule
		\end{tabular}
	\end{table*}
	In this section, we mainly present the experimental results under the IID environment with complete modality input. As can be seen from Table~\ref{ua_iid} and Tabel~\ref{a_iid}, the proposed CIDer maintains good performance on both the CMU-MOSI and CMU-MOSEI datasets. It is important to note that even under the conditions of complete modality input and IID environment, CIDer can still retain the MSSD and the MACI modules, and achieve good performance without any processing, which further demonstrates its excellent robustness when facing different data inputs.
	
	As shown in Table~\ref{cross_a_iid}, when facing domain generalization scenarios, CIDer still achieved the best performance in six metrics under complete modality input on both datasets, even without further special treatment for domain generalization. This further proves the effectiveness of the MACI module in dealing with data distribution shifts.
	
	\section*{Experimental Details of Various Modality Missing Scenarios}
	\begin{table*}[!t]
		\caption{Comparison with state-of-the-art methods.}
		\label{mosi_rmfm_ua_iid}
		\centering
\begin{tabular}{cccccccccccccccc}
			\toprule
			\multirow{4}{*}{Models}  & \multicolumn{15}{c}{CMU-MOSI}                                                                                                   \\
			& \multicolumn{15}{c}{(RMFM/UA/IID)}                                                                                              \\ \cmidrule{2-16} 
			& \multicolumn{3}{c}{0.1} & \multicolumn{3}{c}{0.2} & \multicolumn{3}{c}{0.3} & \multicolumn{3}{c}{0.4} & \multicolumn{3}{c}{0.5} \\ \cmidrule{2-16} 
			& Acc2   & F1     & Acc7  & Acc2   & F1     & Acc7  & Acc2   & F1     & Acc7  & Acc2   & F1     & Acc7  & Acc2   & F1     & Acc7  \\ \midrule
			DMD~\cite{li2023decoupled}                     & 80.8   & 80.9   & \textbf{45.0}  & 79.7   & 79.8   & 40.7  & 78.2   & 78.3   & 37.3  & 73.2   & 73.3   & 32.5  & 71.2   & 71.4   & 33.5  \\
			GLoMo~\cite{zhuang2024glomo}                   & 82.8   & 82.8   & 42.0  & 80.3   & 80.4   & 34.3  & 77.7   & 77.8   & 37.0  & 75.0   & 75.2   & 32.5  & 70.1   & 70.2   & 26.7  \\
			DLF~\cite{wang2024dlf}                     & 81.4   & 81.5   & 41.6  & 79.4   & 79.5   & 37.9  & 75.5   & 75.6   & 35.4  & 73.5   & 73.6   & 33.1  & 69.8   & 70.0   & 28.9  \\ \midrule
			EMT-DLFR~\cite{sun2023efficient}                & 82.2   & 82.2   & 44.0  & \textbf{81.9}   & \textbf{81.8}   & 40.8  & 78.4   & 78.5   & 36.7  & 76.4   & 76.4   & 35.7  & 71.3   & 71.5   & 33.2  \\
\bottomrule
		\end{tabular}
	\end{table*}
	\begin{table*}[!t]
		\caption{Comparison with state-of-the-art methods.}
		\label{mosei_rmfm_ua_iid}
		\centering
		\begin{tabular}{cccccccccccccccc}
			\toprule
			\multirow{4}{*}{Models}  & \multicolumn{15}{c}{CMU-MOSEI}                                                                                                  \\
			& \multicolumn{15}{c}{(RMFM/UA/IID)}                                                                                              \\ \cmidrule{2-16} 
			& \multicolumn{3}{c}{0.1} & \multicolumn{3}{c}{0.2} & \multicolumn{3}{c}{0.3} & \multicolumn{3}{c}{0.4} & \multicolumn{3}{c}{0.5} \\ \cmidrule{2-16} 
			& Acc2   & F1     & Acc7  & Acc2   & F1     & Acc7  & Acc2   & F1     & Acc7  & Acc2   & F1     & Acc7  & Acc2   & F1     & Acc7  \\ \midrule
			DMD~\cite{li2023decoupled}                     & 83.7   & 83.8   & 50.7  & 83.2   & 83.2   & 49.8  & 81.7   & 81.6   & 49.1  & 78.5   & 78.7   & 46.8  & 76.5   & 76.4   & 45.2  \\
			GLoMo~\cite{zhuang2024glomo}                   & \textbf{85.1}   & \textbf{85.0}   & 51.7  & \textbf{83.6}   & \textbf{83.4}   & 49.4  & 81.6   & 81.3   & 47.2  & \textbf{80.4}   & 79.9   & 46.6  & 77.8   & 77.3   & 44.6  \\
			DLF~\cite{wang2024dlf}                     & 83.7   & 83.8   & 51.1  & 82.5   & 82.4   & 48.7  & 80.8   & 80.7   & 46.3  & 78.2   & 78.1   & 47.5  & 77.1   & 77.1   & 45.5  \\ \midrule
			EMT-DLFR~\cite{sun2023efficient}                & 82.4   & 82.3   & 51.6  & 80.9   & 81.0   & 51.2  & 79.4   & 79.3   & 48.1  & 75.4   & 75.3   & 45.7  & 76.1   & 75.7   & 45.6  \\
\bottomrule
		\end{tabular}
	\end{table*}
	\begin{table*}[!t]
		\caption{Comparison with state-of-the-art methods.}
		\label{traditional_rmfm_ua_iid}
		\centering
		\begin{tabular}{cccccccccccccccc}
			\toprule
			\multirow{4}{*}{Models}  & \multicolumn{15}{c}{CMU-MOSI}                                                                                                   \\
			& \multicolumn{15}{c}{(Traditional RMFM/UA/IID)}                                                                                  \\ \cmidrule{2-16} 
			& \multicolumn{3}{c}{0.1} & \multicolumn{3}{c}{0.2} & \multicolumn{3}{c}{0.3} & \multicolumn{3}{c}{0.4} & \multicolumn{3}{c}{0.5} \\ \cmidrule{2-16} 
			& Acc2   & F1     & Acc7  & Acc2   & F1     & Acc7  & Acc2   & F1     & Acc7  & Acc2   & F1     & Acc7  & Acc2   & F1     & Acc7  \\ \midrule
			DMD~\cite{li2023decoupled}                     & \textbf{82.8}   & \textbf{82.8}   & 41.6  & 79.1   & 79.2   & 40.2  & 77.0   & 77.1   & \textbf{40.8}  & 74.9   & 75.0   & \textbf{36.0}  & 68.1   & 68.2   & 28.0  \\
			GLoMo~\cite{zhuang2024glomo}                   & 82.0   & 81.9   & \textbf{45.3}  & 77.7   & 77.8   & 39.4  & 76.8   & 76.9   & 38.5  & 72.1   & 72.3   & 26.8  & 71.3   & 71.5   & 26.7  \\
			DLF~\cite{wang2024dlf}                     & 80.6   & 80.7   & 41.6  & 80.0   & 80.1   & 39.7  & \textbf{78.2}   & \textbf{78.3}   & 36.6  & 75.6   & 75.6   & 33.5  & 70.3   & 70.4   & 30.5  \\ \midrule
			EMT-DLFR~\cite{sun2023efficient}                & 82.2   & 82.2   & 42.6  & 78.8   & 78.9   & 39.1  & 77.7   & 77.9   & 38.5  & 73.9   & 74.1   & 34.6  & 71.8   & 72.0   & \textbf{34.3}  \\
\bottomrule
		\end{tabular}
	\end{table*}
	\begin{table*}[!t]
		\caption{Comparison with state-of-the-art methods.}
		\label{rmm_ua_iid}
		\centering
		\begin{tabular}{cccccccccccccccc}
			\toprule
			\multirow{4}{*}{Models} & \multicolumn{15}{c}{CMU-MOSI}                                                                                                   \\
			& \multicolumn{15}{c}{(RMM/UA/IID)}                                                                                               \\ \cmidrule{2-16} 
			& \multicolumn{3}{c}{0.1} & \multicolumn{3}{c}{0.2} & \multicolumn{3}{c}{0.3} & \multicolumn{3}{c}{0.4} & \multicolumn{3}{c}{0.5} \\ \cmidrule{2-16} 
			& Acc2   & F1     & Acc7  & Acc2   & F1     & Acc7  & Acc2   & F1     & Acc7  & Acc2   & F1     & Acc7  & Acc2   & F1     & Acc7  \\ \midrule
			DMD~\cite{li2023decoupled}                     & 81.6   & 81.5   & 43.9  & 76.1   & 76.2   & 37.8  & 72.6   & 72.6   & 36.3  & 73.9   & 74.0   & 34.6  & 66.8   & 66.9   & 26.4  \\
			GLoMo~\cite{zhuang2024glomo}                   & 82.2   & 82.1   & 43.9  & 77.7   & 77.9   & 40.5  & 72.7   & 72.9   & 34.6  & 68.8   & 68.9   & 27.1  & 65.2   & 65.0   & 28.3  \\
			DLF~\cite{wang2024dlf}                     & 81.3   & 81.3   & \textbf{44.6}  & 76.5   & 76.7   & 38.2  & 73.5   & 73.6   & 37.0  & 69.4   & 69.4   & 33.8  & 60.2   & 58.0   & 27.7  \\ \midrule
			EMT-DLFR~\cite{sun2023efficient}                & 81.7   & 81.7   & 43.4  & 76.5   & 76.7   & 38.6  & 75.3   & 75.4   & 37.9  & \textbf{76.7}   & \textbf{76.5}   & 35.9  & 68.5   & 68.6   & 29.7  \\
\bottomrule
		\end{tabular}
	\end{table*}
	\begin{table*}[!t]
		\caption{Comparison with state-of-the-art methods.}
		\label{mosi_rmfm_ua_ood}
		\centering
		\begin{tabular}{cccccccccccccccc}
			\toprule
			\multirow{4}{*}{Models} & \multicolumn{15}{c}{CMU-MOSI}                                                                                                   \\
			& \multicolumn{15}{c}{(RMFM/UA/OOD)}                                                                                              \\ \cmidrule{2-16} 
			& \multicolumn{3}{c}{0.1} & \multicolumn{3}{c}{0.2} & \multicolumn{3}{c}{0.3} & \multicolumn{3}{c}{0.4} & \multicolumn{3}{c}{0.5} \\ \cmidrule{2-16} 
			& Acc2   & F1     & Acc7  & Acc2   & F1     & Acc7  & Acc2   & F1     & Acc7  & Acc2   & F1     & Acc7  & Acc2   & F1     & Acc7  \\ \midrule
			DMD~\cite{li2023decoupled}                     & 70.3   & 70.3   & \textbf{41.6}  & 69.1   & 69.3   & 36.4  & 60.6   & 60.4   & 33.2  & 62.4   & 62.6   & 28.3  & 58.0   & 58.1   & 25.2  \\
			GLoMo~\cite{zhuang2024glomo}                   & 71.0   & 71.1   & 40.2  & 71.9   & 71.8   & 32.9  & 62.1   & 62.2   & 34.3  & 63.2   & 63.2   & 32.7  & 57.7   & 57.7   & 21.3  \\
			DLF~\cite{wang2024dlf}                     & 69.0   & 69.1   & 39.1  & 66.7   & 66.7   & 33.1  & 62.4   & 62.5   & 29.5  & 61.8   & 61.9   & 29.7  & 59.6   & 59.8   & 22.6  \\ \midrule
			EMT-DLFR~\cite{sun2023efficient}                & 72.0   & 72.1   & 36.6  & 69.0   & 69.1   & 34.8  & 66.5   & 66.6   & 35.0  & 63.5   & 63.6   & 33.4  & 63.9   & 64.1   & 30.5  \\
\bottomrule
		\end{tabular}
	\end{table*}
	\begin{table*}[!t]
		\caption{Comparison with state-of-the-art methods.}
		\label{mosei_rmfm_ua_ood}
		\centering
		\begin{tabular}{cccccccccccccccc}
			\toprule
			\multirow{4}{*}{Models} & \multicolumn{15}{c}{CMU-MOSEI}                                                                                                  \\
			& \multicolumn{15}{c}{(RMFM/UA/OOD)}                                                                                              \\ \cmidrule{2-16} 
			& \multicolumn{3}{c}{0.1} & \multicolumn{3}{c}{0.2} & \multicolumn{3}{c}{0.3} & \multicolumn{3}{c}{0.4} & \multicolumn{3}{c}{0.5} \\ \cmidrule{2-16} 
			& Acc2   & F1     & Acc7  & Acc2   & F1     & Acc7  & Acc2   & F1     & Acc7  & Acc2   & F1     & Acc7  & Acc2   & F1     & Acc7  \\ \midrule
			DMD~\cite{li2023decoupled}                     & 69.7   & 70.0   & 47.8  & 70.2   & 70.5   & 46.0  & 65.4   & 65.3   & 44.3  & 61.2   & 60.6   & 41.9  & 62.4   & 62.4   & 42.5  \\
			GLoMo~\cite{zhuang2024glomo}                   & 71.3   & 71.5   & \textbf{48.4}  & 68.1   & 68.3   & \textbf{47.2}  & 68.0   & 68.2   & 45.6  & 64.0   & 64.1   & 42.4  & 62.7   & 62.6   & 41.9  \\
			DLF~\cite{wang2024dlf}                     & 70.0   & 70.1   & 46.8  & 69.5   & 69.8   & 45.2  & 66.8   & 67.0   & 44.7  & 64.5   & 64.8   & 42.8  & 62.3   & 62.5   & 42.4  \\ \midrule
			EMT-DLFR~\cite{sun2023efficient}                & 69.9   & 70.2   & 48.0  & 67.0   & 67.1   & 46.6  & 66.7   & 66.8   & 44.6  & 61.4   & 61.3   & 43.8  & 60.5   & 60.3   & 42.5  \\
\bottomrule
		\end{tabular}
	\end{table*}
	\begin{table*}[!t]
		\caption{Comparison with state-of-the-art methods.}
		\label{mosi_cross_rmfm_a_iid}
		\centering
		\begin{tabular}{cccccccccccccccc}
			\toprule
			\multirow{4}{*}{Models} & \multicolumn{15}{c}{CMU-MOSI $\rightarrow$ CMU-MOSEI}                                                                                       \\
			& \multicolumn{15}{c}{(RMFM/A/IID)}                                                                                               \\ \cmidrule{2-16} 
			& \multicolumn{3}{c}{0.1} & \multicolumn{3}{c}{0.2} & \multicolumn{3}{c}{0.3} & \multicolumn{3}{c}{0.4} & \multicolumn{3}{c}{0.5} \\ \cmidrule{2-16} 
			& Acc2   & F1     & Acc7  & Acc2   & F1     & Acc7  & Acc2   & F1     & Acc7  & Acc2   & F1     & Acc7  & Acc2   & F1     & Acc7  \\ \midrule
			GLoMo~\cite{zhuang2024glomo}                   & \textbf{79.6}   & \textbf{79.7}   & 41.7  & 76.8   & 76.9   & 41.0  & 72.5   & 72.9   & 38.1  & 70.2   & 70.7   & 36.3  & 68.2   & 68.4   & 33.2  \\ \midrule
			EMT-DLFR~\cite{sun2023efficient}                & 75.8   & 75.9   & 38.5  & 75.7   & 75.9   & 40.0  & 70.2   & 70.7   & 35.7  & 49.9   & 48.5   & 23.7  & 61.6   & 62.1   & 30.1  \\
			LNLN~\cite{zhang2024towards}                    & 78.3   & 78.6   & 37.5  & 76.4   & 76.9   & 36.2  & 75.0   & \textbf{75.7}   & 34.8  & \textbf{73.4}   & \textbf{74.2}   & 33.2  & \textbf{71.8}   & \textbf{72.9}   & 32.1  \\ \midrule
			DiCMoR~\cite{wang2023distribution}                  & 78.9   & 78.9   & 36.3  & \textbf{77.5}   & \textbf{77.1}   & 32.9  & \textbf{75.6}   & 75.3   & 28.8  & 73.1   & 72.7   & 30.9  & 71.1   & 69.7   & 30.0  \\
\bottomrule
		\end{tabular}
	\end{table*}
	\begin{table*}[!t]
		\caption{Comparison with state-of-the-art methods.}
		\label{mosei_cross_rmfm_a_iid}
		\centering
		\begin{tabular}{cccccccccccccccc}
			\toprule
			\multirow{4}{*}{Models} & \multicolumn{15}{c}{CMU-MOSEI → CMU-MOSI}                                                                                       \\
			& \multicolumn{15}{c}{(RMFM/A/IID)}                                                                                               \\ \cmidrule{2-16} 
			& \multicolumn{3}{c}{0.1} & \multicolumn{3}{c}{0.2} & \multicolumn{3}{c}{0.3} & \multicolumn{3}{c}{0.4} & \multicolumn{3}{c}{0.5} \\ \cmidrule{2-16} 
			& Acc2   & F1     & Acc7  & Acc2   & F1     & Acc7  & Acc2   & F1     & Acc7  & Acc2   & F1     & Acc7  & Acc2   & F1     & Acc7  \\ \midrule
			GLoMo~\cite{zhuang2024glomo}                   & 79.7   & 79.6   & 36.4  & 78.6   & 78.4   & 35.7  & 76.7   & 76.6   & 36.7  & 73.3   & 73.4   & 32.0  & 67.3   & 67.5   & 27.4  \\ \midrule
			EMT-DLFR~\cite{sun2023efficient}                & 80.1   & 80.1   & 38.6  & 77.0   & 77.2   & 32.3  & 74.7   & 74.7   & 33.9  & 67.0   & 67.2   & 27.7  & 63.5   & 63.6   & 24.6  \\
			LNLN~\cite{zhang2024towards}                    & 80.1   & 80.2   & 41.2  & 78.1   & 78.1   & 40.3  & 75.8   & 75.8   & 36.0  & 71.2   & 71.1   & 32.0  & 69.0   & 69.0   & 29.7  \\ \midrule
			DiCMoR~\cite{wang2023distribution}                  & 75.8   & 75.7   & 36.6  & 74.1   & 74.2   & 35.0  & 71.3   & 71.4   & 33.0  & 68.0   & 68.2   & 29.4  & 65.2   & 65.4   & 24.5  \\
\bottomrule
		\end{tabular}
	\end{table*}
	\begin{table*}[!t]
		\caption{Comparison with state-of-the-art methods.}
		\label{tmfm_a_iid}
		\centering
		\begin{tabular}{cccccccccccccccc}
			\toprule
			\multirow{4}{*}{Models} & \multicolumn{15}{c}{CMU-MOSI}                                                                                                   \\
			& \multicolumn{15}{c}{(TMFM/A/IID)}                                                                                               \\ \cmidrule{2-16} 
			& \multicolumn{3}{c}{0.1} & \multicolumn{3}{c}{0.2} & \multicolumn{3}{c}{0.3} & \multicolumn{3}{c}{0.4} & \multicolumn{3}{c}{0.5} \\ \cmidrule{2-16} 
			& Acc2   & F1     & Acc7  & Acc2   & F1     & Acc7  & Acc2   & F1     & Acc7  & Acc2   & F1     & Acc7  & Acc2   & F1     & Acc7  \\ \midrule
			DMD~\cite{li2023decoupled}                     & 80.2   & 80.3   & 41.8  & 79.1   & 79.2   & 40.7  & 76.7   & 76.8   & 39.5  & \textbf{76.1}   & \textbf{76.2}   & 37.2  & 71.0   & 71.2   & 29.6  \\
			GLoMo~\cite{zhuang2024glomo}                   & 81.7   & 81.8   & 40.5  & 78.7   & 78.8   & 39.9  & 74.9   & 75.0   & 34.7  & 74.4   & 74.5   & 31.3  & 68.3   & 68.3   & 25.4  \\
			DLF~\cite{wang2024dlf}                     & 80.2   & 80.3   & 40.1  & 78.7   & 78.8   & 39.4  & 75.5   & 75.6   & 34.7  & 72.0   & 72.1   & 29.5  & 66.3   & 65.7   & 26.1  \\ \midrule
			EMT-DLFR~\cite{sun2023efficient}                & 82.2   & 82.2   & 42.7  & \textbf{81.4}   & \textbf{81.4}   & \textbf{42.0}  & \textbf{78.2}   & \textbf{78.3}   & \textbf{39.9}  & 74.7   & 74.8   & 36.7  & 72.7   & \textbf{72.9}   & \textbf{34.0}  \\
\bottomrule
		\end{tabular}
	\end{table*}
	\begin{table*}[!t]
		\caption{Comparison with state-of-the-art methods.}
		\label{stmfm_a_iid}
		\centering
		\begin{tabular}{cccccccccccccccc}
			\toprule
			\multirow{4}{*}{Models} & \multicolumn{15}{c}{CMU-MOSI}                                                                                                   \\
			& \multicolumn{15}{c}{(STMFM/A/IID)}                                                                                              \\ \cmidrule{2-16} 
			& \multicolumn{3}{c}{0.1} & \multicolumn{3}{c}{0.2} & \multicolumn{3}{c}{0.3} & \multicolumn{3}{c}{0.4} & \multicolumn{3}{c}{0.5} \\ \cmidrule{2-16} 
			& Acc2   & F1     & Acc7  & Acc2   & F1     & Acc7  & Acc2   & F1     & Acc7  & Acc2   & F1     & Acc7  & Acc2   & F1     & Acc7  \\ \midrule
			DMD~\cite{li2023decoupled}                     & \textbf{83.7}   & \textbf{83.8}   & \textbf{46.4}  & 79.0   & 79.1   & 36.6  & 75.0   & 75.2   & 35.4  & 74.1   & 74.2   & 33.8  & 70.9   & 71.1   & 30.5  \\
			GLoMo~\cite{zhuang2024glomo}                   & \textbf{83.7}   & 83.7   & 44.2  & \textbf{81.1}   & \textbf{81.2}   & 41.3  & 77.9   & 78.0   & 36.0  & 73.8   & 73.9   & \textbf{35.7}  & 72.4   & 72.6   & 32.9  \\
			DLF~\cite{wang2024dlf}                     & 82.9   & 83.0   & 40.5  & 78.8   & 78.9   & 37.9  & 75.9   & 76.0   & 35.3  & 73.2   & 73.3   & 34.4  & 72.0   & 72.1   & 29.0  \\ \midrule
			EMT-DLFR~\cite{sun2023efficient}                & 81.1   & 81.2   & 42.9  & 80.8   & 80.8   & 38.8  & \textbf{78.5}   & \textbf{78.6}   & 35.7  & \textbf{75.5}   & \textbf{75.6}   & 35.4  & \textbf{74.7}   & \textbf{74.8}   & \textbf{33.2}  \\
\bottomrule
		\end{tabular}
	\end{table*}
	To more clearly present the performance of CIDer and various state-of-the-art methods under different modality missing scenarios and different missing rates, we report the specific numerical results in tabular form in this section.
	
	\begin{itemize}
		\item \textbf{RMFM/UA/IID:} Table~\ref{mosi_rmfm_ua_iid} and Table~\ref{mosei_rmfm_ua_iid}.
		\item \textbf{Traditional RMFM/UA/IID:} Table~\ref{traditional_rmfm_ua_iid}.
		\item \textbf{RMM/UA/IID:} Table~\ref{rmm_ua_iid}.
		\item \textbf{RMFM/UA/OOD:} Table~\ref{mosi_rmfm_ua_ood} and Table~\ref{mosei_rmfm_ua_ood}.
		\item \textbf{[Cross-dataset] RMFM/A/IID:} Table~\ref{mosi_cross_rmfm_a_iid} and Table~\ref{mosei_cross_rmfm_a_iid}.
		\item \textbf{TMFM/A/IID:} Table~\ref{tmfm_a_iid}.
		\item \textbf{STMFM/A/IID:} Table~\ref{stmfm_a_iid}.
	\end{itemize}
}

\bibliographystyle{IEEEtran}
\bibliography{IEEEexample}

\end{document}